\documentclass[10pt,journal,compsoc]{IEEEtran}

\ifCLASSOPTIONcompsoc
  \usepackage[nocompress]{cite}
\else
  \usepackage{cite}
\fi
\ifCLASSINFOpdf
\else
\fi
\usepackage{amsmath}
\usepackage{amssymb}
\usepackage{multirow}
\usepackage{array}
\usepackage{placeins}
\usepackage{amsmath,bm}
\usepackage{array}
\usepackage{amsmath}
\usepackage{amsthm}
\usepackage{amsfonts}
\usepackage{booktabs}
\usepackage{algorithm}
\usepackage{algorithmic}

\usepackage{xcolor}
\usepackage{graphicx}  
\usepackage{dsfont}

\usepackage[pagebackref=true,breaklinks=true,colorlinks,bookmarks=false,linkcolor=blue,citecolor=blue,urlcolor=blue]{hyperref}

\usepackage{url}
\usepackage{enumerate}
\usepackage{multirow}
\usepackage{booktabs}  
\usepackage{makecell}
\usepackage{bm}
\usepackage{pifont}
\usepackage{threeparttable} 
\usepackage{ragged2e}  
\usepackage[edges]{forest}
\usepackage{tikz}
\usepackage{verbatim}
\usepackage{multicol}
\usepackage[numbers]{natbib}
\usepackage{tabularx}
\usepackage{subfigure}

\usepackage{colortbl}

\definecolor{mygreen}{RGB}{166,219,162}
\definecolor{mypurple}{RGB}{0,153,204}
\definecolor{output-black}{RGB}{122,122,122}
\definecolor{myblue_1}{RGB}{230, 255, 255}
\definecolor{mypink_1}{RGB}{255, 240, 255}
\definecolor{mycolor_0}{RGB}{210, 230, 222}
\definecolor{mycolor_1}{RGB}{207, 245, 245}
\definecolor{mycolor_2}{RGB}{190, 220, 230}

\definecolor{mycolor_3}{RGB}{214, 229, 247}
\definecolor{mycolor_4}{RGB}{197, 235, 245}

\definecolor{mycolor_5}{RGB}{243, 238, 223}
\definecolor{mycolor_6}{RGB}{240, 235, 255}
\definecolor{mycolor_7}{RGB}{210, 240, 230}
\definecolor{mycolor_8}{RGB}{180, 220, 250}

\definecolor{mytabcolor_1}{RGB}{234, 250, 245}
\definecolor{mytabcolor_2}{RGB}{249, 246, 237}
\definecolor{mytabcolor_3}{RGB}{246, 243, 255}
\definecolor{mytabcolor_4}{RGB}{236, 248, 250}

\definecolor{revision-color_1}{RGB}{0, 0, 0}

\begin{document}
%
\title{CLIP-Powered Domain Generalization and Domain Adaptation: A Comprehensive Survey}
%
%
%
%

\author{
        Jindong Li,
        Yongguang Li,
        Yali Fu, 
        Jiahong Liu,
        Yixin Liu,
        Menglin Yang*,
        Irwin King~\IEEEmembership{Fellow,~IEEE}
        
\IEEEcompsocitemizethanks{
\IEEEcompsocthanksitem Jindong Li and Menglin Yang are with The Hong Kong University of Science and Technology (Guangzhou), Guangzhou, China. E-mail: jli839@connect.hkust-gz.edu.cn, menglinyang@hkust-gz.edu.cn.
\IEEEcompsocthanksitem Yongguang Li and Yali Fu are with The Jilin University, Changchun, China. E-mail: \{ygli25, fuyl23\}@mails.jlu.edu.cn.
\IEEEcompsocthanksitem Yixin Liu is with Griffith University, Queensland, Australia. E-mail: Yixin.liu@griffith.edu.au.
\IEEEcompsocthanksitem Jiahong Liu and Irwin King are with The Chinese University of Hong Kong, Hong Kong, China. E-mail: jiahong.liu21@gmail.com, king@cse.cuhk.edu.hk.
\IEEEcompsocthanksitem *: Corresponding author.
}
\thanks{Manuscript received xxx xx, 20xx; revised xxxx xx, 20xx.}}

%
%

\markboth{Journal of \LaTeX\ Class Files,~Vol.~14, No.~8, August~2015}%
{Shell \MakeLowercase{\textit{et al.}}: Bare Demo of IEEEtran.cls for Computer Society Journals}
%



\IEEEtitleabstractindextext{%
\begin{abstract}
\justifying
\textcolor{revision-color_1}{
As machine learning evolves, domain generalization (DG) and domain adaptation (DA) have become crucial for improving model robustness across diverse environments. Contrastive Language–Image Pretraining (CLIP) plays a central role in these tasks, offering strong zero-shot capabilities that allow models to operate effectively in unseen domains. Yet, despite CLIP’s growing influence, no comprehensive survey has systematically examined its applications in DG and DA, underscoring the need for this review.
This survey provides a unified and in-depth overview of CLIP-driven DG and DA. 
Before reviewing methods, we establish precise and complete scenario definitions covering source accessibility (SA vs. SF), source number (SS vs. MS), and label relations (CS, PS, OS, OPS), forming a coherent taxonomy that structures all subsequent analyses. 
For DG, we categorize methods into prompt optimization techniques that enhance task alignment and architectures that leverage CLIP as a backbone for transferable feature extraction. 
For DA, we examine both source-available approaches that rely on labeled source data and source-free approaches operating primarily on target-domain samples, emphasizing the knowledge transfer mechanisms that enable adaptation across heterogeneous settings. 
We further provide consolidated trend analyses for both DG and DA, revealing overarching patterns, methodological principles, and scenario-dependent behaviors.
We then discuss key challenges such as realistic deployment scenarios, LLM knowledge integration, multimodal fusion, interpretability, and catastrophic forgetting, and outline future directions for developing scalable and trustworthy CLIP-based DG and DA systems. 
By synthesizing existing studies and highlighting critical gaps, this survey offers actionable insights for researchers and practitioners, motivating new strategies for leveraging CLIP to advance domain robustness in real-world scenarios.
A continuously updated list of related works is maintained at: \url{https://github.com/jindongli-Ai/Survey_on_CLIP-Powered_Domain_Generalization_and_Adaptation}.
}
\end{abstract}

\begin{IEEEkeywords}
Vision-Language Model (VLM), CLIP-Powered, Domain Generalization (DG), Domain Adaptation (DA).
\end{IEEEkeywords}}

\maketitle

\IEEEdisplaynontitleabstractindextext

%
\IEEEpeerreviewmaketitle

\IEEEraisesectionheading{
\section{Introduction}
\label{sec_Introduction}}


%
%
%
%

 

\IEEEPARstart{A}{s} machine learning progresses, \textcolor{revision-color_1}{domain generalization (DG) \cite{2023_ICCV_BorLan, 2023_ICML_CLIPood, 2024_CVPR_MMA, 2025_CVPR_FrogDogNet_FrogDogNet=Fourier-Frequency-Retained-Visual-Prompt-Output-Guidance-for-Domain-Generalization-of-CLIP-in-Remote-Sensing, 2025_TMLR_MetaPrompt_Meta-Learning-to-Teach-Semantic-Prompts-for-Open-Domain-Generalization-in-Vision-Language-Models, 2025_TCSVT_ClipMix_ClipMix-for-Domain-Generalization} and domain adaptation (DA) \cite{2023_ICCV_PADCLIP, 2023_NeurIPS_MPA, 2024_CVPR_DAMP, 2024_ICML_UEO, 2025_ICLR_ProDe_Proxy-Denoising-for-Source-Free-Domain-Adaptation, 2025_TOMM_UTISA_Unified-Text-image-Space-Alignment-with-Cross-Modal-Prompting-in-CLIP-for-UDA} have emerged as pivotal areas of research aimed at enhancing model robustness across diverse environments~\cite{2024_TPAMI_Survey_A-Comprehensive-Survey-on-Source-Free-Domain-Adaptation,2024_IJCV_Survey_A-Comprehensive-Survey-on-Test-time-Adaptation-under-Distribution-Shifts}.} Traditional models often rely on extensive labeled data, which can be challenging to obtain due to factors such as time, cost, and privacy concerns. Moreover, the scarcity of labeled datasets, combined with distribution discrepancies between source and target domains, significantly impedes the generalization capabilities of a model. These challenges highlight the urgent need for robust techniques that can adapt to unseen domains or variations in data distribution. Therefore, developing effective DG and DA methods is crucial for a wide array of applications, particularly in fields like \textcolor{revision-color_1}{general object recognition \cite{2017_CVPR_Office-Home-dataset_Deep-Hashing-Network-for-Unsupervised-Domain-Adaptation, 2010_ECCV_Office-31-dataset_Adapting-Visual-Category-Models-to-New-Domains, 2017_ICCV_PACS-dataset_Deep-Broader-and-Artier-Domain-Generalization, 2013_ICCV_VLCS-dataset_Unbiased-Metric-Learning--On-the-Utilization-of-Multiple-Datasets-and-Web-Images-for-Softening-Bias}, scene recognition \cite{2018_ECCV_Recognition-in-Terra-Incognita, 2023_CVPR_NICO++}, remote sensing \cite{2025_CVPR_FrogDogNet_FrogDogNet=Fourier-Frequency-Retained-Visual-Prompt-Output-Guidance-for-Domain-Generalization-of-CLIP-in-Remote-Sensing, 2025_TGRS_PADA-Net_Prompt-Integrated-Adversarial-Unsupervised-Domain-Adaptation-for-Scene-Recognition},} where models must perform reliably in dynamic and unpredictable environments.

\begin{figure}[ht]
    \centering
    \includegraphics[width=0.99\linewidth]{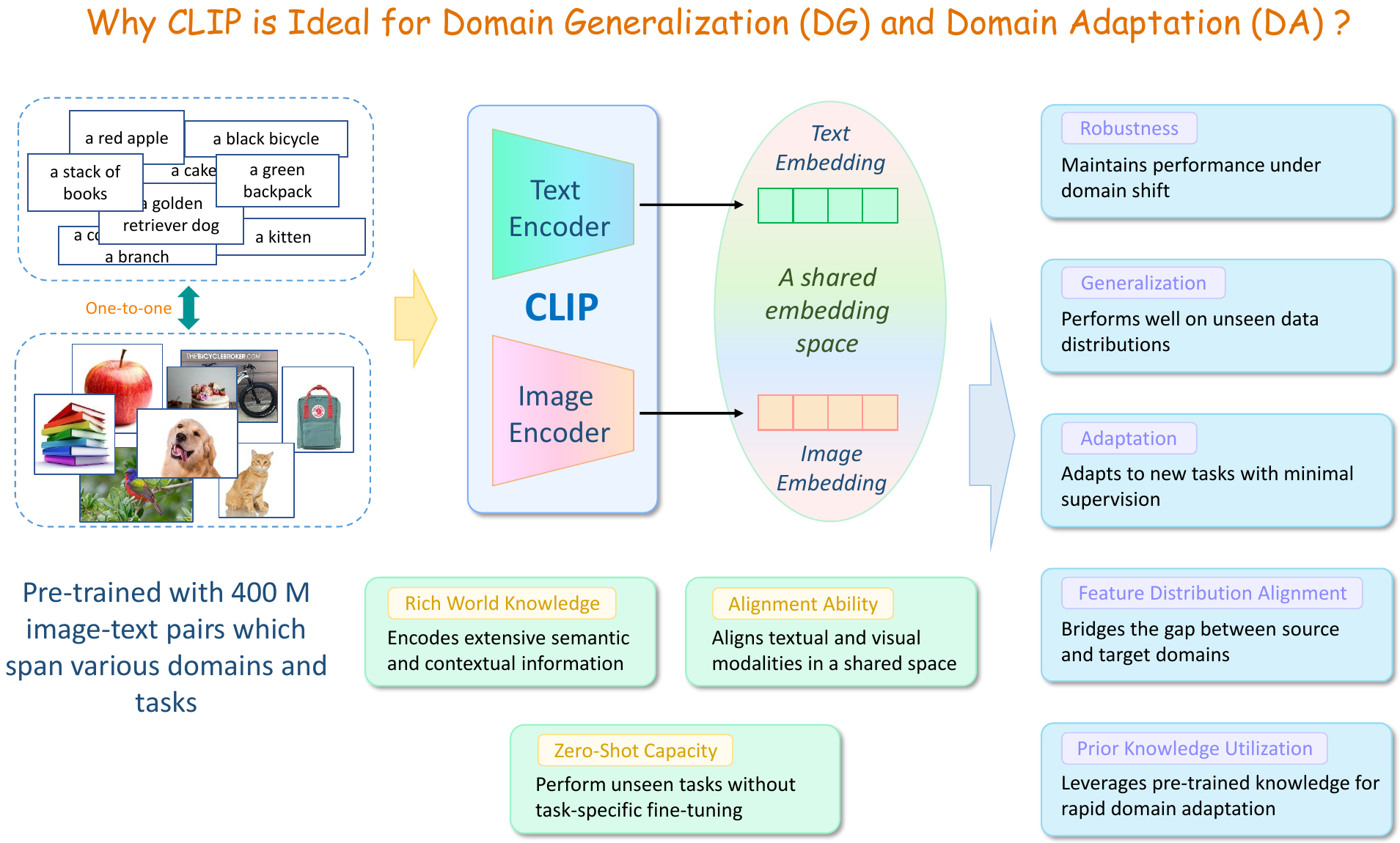}
    \caption{\textcolor{revision-color_1}{The characteristics of CLIP and its perfect fit for Domain Generalization (DG) and Domain Adaptation (DA).}}
    \label{fig:fig1}
\end{figure}

\begin{figure*}[t!]
    \centering
    \resizebox{0.98\linewidth}{!}{
        \input{figs/fig_tree}
        }
    \caption{\textcolor{revision-color_1}{Structure of this paper with representative works, mainly categorized by source availability and label space scenarios (CS, PS, OS, OPS).}}
    \label{fig:fig_tree}
\end{figure*}

Inspired by earlier works such as VirTex~\cite{2021_CVPR_VirTex} and ICMLM~\cite{2020_ECCV_ICMLM}, which demonstrated the effectiveness of transformer-based language modeling, masked language modeling, and contrastive objectives for learning image representations from text, CLIP~\cite{2021_ICML_CLIP} adopts a contrastive learning framework. By leveraging 400 million image–text pairs, CLIP learns rich multi-modal representations that align semantically related images and texts while distinguishing irrelevant ones. This alignment captures intricate visual–linguistic relationships and enhances robustness to distributional variations. 
Moreover, CLIP supports flexible task adaptation through prompting~\cite{2023_ACM-Computring-Surveys_Survey_Pre-train-Prompt-and-Predict-A-Systematic-Survey-of-Prompting-Methods-in-Natural-Language-Processing, 2023_arXiv_Survey_A-Systematic-Survey-of-Prompt-Engineering-on-Vision-Language-Foundation-Models, 2024_arXiv_Survey_The-Prompt-Report--A-Systematic-Survey-of-Prompting-Techniques}, enabling contextual interpretation of visual data.
As illustrated in Fig.~\ref{fig:fig1}, these characteristics make CLIP particularly effective for DG and DA. Its powerful zero-shot capabilities allow models to perform in unseen domains without task-specific fine-tuning, substantially advancing cross-domain generalization and reshaping the research landscape of DG and DA.

\begin{figure*}[t]
    \centering
    \includegraphics[width=0.98\linewidth]{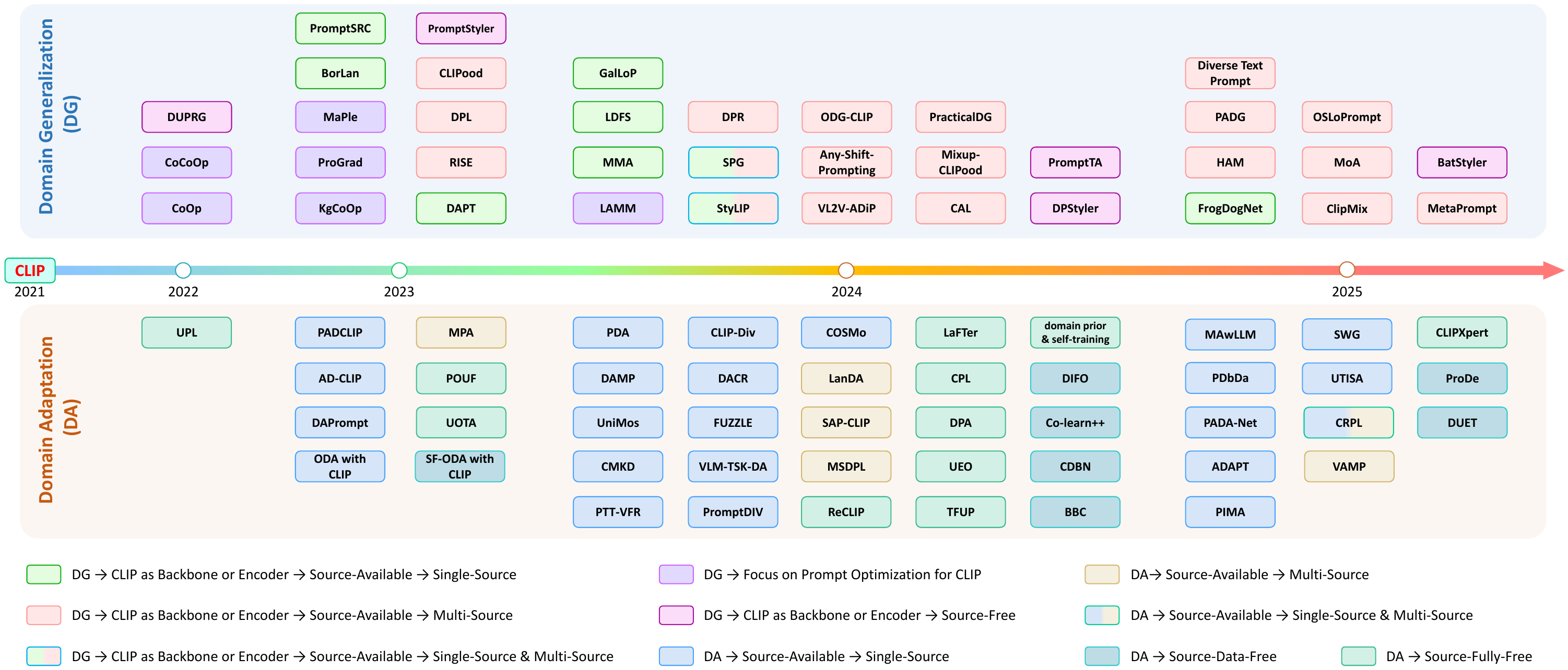}
    \caption{\textcolor{revision-color_1}{Roadmap (i.e., timeline) of CLIP-powered Domain Generalization (DG) and Domain Adaptation (DA). The research activity peaked in 2024, showing rapid progress in both fields. CLIP-based DA methods trend from Source-Available (SA) to Source-Free (SF), while DG methods evolve from Single-Source (SS) to Multi-Source (MS) settings.}}
    \label{fig:Roadmap}
\end{figure*}

Given its remarkable suitability for DG and DA tasks, CLIP has garnered increasing attention in recent years. Researchers have actively explored CLIP-powered methods, proposing innovative approaches to address domain shifts and enhance cross-domain performance. As illustrated in Fig.~\ref{fig:Roadmap}, the number of works leveraging CLIP for DG and DA has surged rapidly, reflecting its growing prominence in the field. This trend underscores the importance of systematically understanding and categorizing these methodologies to fully unlock the potential of CLIP in DG and DA.

Building on this growing research interest, several surveys have attempted to summarize progress in domain generalization and adaptation~\cite{2022_TKDE_Survey_Generalizing-to-Unseen-Domains=A-Survey-on-Domain-Generalization, 2024_AIReview_Survey_Domain-Generalization-through-Meta-learning=A-Survey, 2024_TPAMI_Survey_A-Comprehensive-Survey-on-Source-Free-Domain-Adaptation, 2024_IJCV_Survey_A-Comprehensive-Survey-on-Test-time-Adaptation-under-Distribution-Shifts, 2024_NN_Survey_Source-free-Unsupervised-Domain-Adaptation=A-Survey}.
Existing DG surveys~\cite{2022_TKDE_Survey_Generalizing-to-Unseen-Domains=A-Survey-on-Domain-Generalization, 2024_AIReview_Survey_Domain-Generalization-through-Meta-learning=A-Survey} provide comprehensive overviews of generalization theories, meta-learning strategies, and domain-invariant learning. However, their analyses mainly revolve around vision-only architectures and do not systematically examine multi-modal alignment or prompt-driven mechanisms that are essential to CLIP-based approaches.
In the adaptation literature, SFDA surveys~\cite{2024_TPAMI_Survey_A-Comprehensive-Survey-on-Source-Free-Domain-Adaptation, 2024_NN_Survey_Source-free-Unsupervised-Domain-Adaptation=A-Survey} concentrate on source-free adaptation methods built on unimodal visual backbones, with emphasis on pseudo-labeling, adversarial optimization, and self-training.
Likewise, the TTA survey~\cite{2024_IJCV_Survey_A-Comprehensive-Survey-on-Test-time-Adaptation-under-Distribution-Shifts} reviews inference-time parameter adaptation strategies, yet still within traditional single-modality visual pipelines.
Overall, these surveys are grounded in conventional vision-only frameworks and therefore do not examine CLIP’s essential mechanisms such as large-scale vision–language pretraining, multi-modal alignment, and prompt-based task adaptation.

To bridge this gap, this survey provides a systematic consolidation of recent advances in CLIP-powered domain generalization (DG) and domain adaptation (DA). We analyze representative methodologies, summarize practical strategies for leveraging CLIP across diverse scenarios, and offer a unified perspective on how its multi-modal representations and zero-shot capabilities contribute to cross-domain robustness, as illustrated in Fig.~\ref{fig:fig_tree}.

Specifically, we introduce the theoretical foundations in (\S\ref{sec_Preliminaries}), conduct a systematic review of CLIP-based DG (\S\ref{sec_DG}) and DA (\S\ref{sec_DA}) methods organized by source availability (i.e., source-available (SA), source-free (SF)) and label-space scenarios (i.e., closed-set (CS), partial-set (PS), open-set (OS), open-partial-set (OPS)), and summarize comprehensive benchmarks and evaluation metrics in (\S\ref{sec_Benchmarks-and-Metrics}).
We further discuss emerging challenges and future directions in (\S\ref{sec_Challenges-and-Future-Directions}), including realistic deployment scenarios, LLM knowledge incorporation, multimodal integration, interpretability, and catastrophic forgetting. Finally, we conclude the survey in (\S\ref{sec_Conclusion}) with a concise summary of insights and key takeaways from CLIP-powered domain generalization and adaptation.

The contributions of this survey could be summarized as follows: 
\begin{itemize}
    \item This survey provides the first comprehensive review of CLIP-powered domain generalization (DG) and domain adaptation (DA), establishing a unified taxonomy that organizes existing methods and highlights CLIP’s advantages in multi-modal representation learning and cross-domain transfer.
    \item We conduct comparative analyses of representative DG and DA approaches under both source-available and source-free settings, accompanied by discussions on benchmark performance and method characteristics that offer insights for developing more generalizable CLIP-based frameworks.
    \item We provide an overview of datasets and evaluation metrics commonly used in DG and DA research. This overview serves as a practical reference for selecting benchmarks and metrics to evaluate CLIP-based models across diverse domain shifts.
    \item We provide an in-depth discussion of key challenges and future directions, highlighting how CLIP’s unique capabilities can be leveraged to overcome current limitations and drive further progress in DG and DA.
\end{itemize}

\section{Preliminaries} 
\label{sec_Preliminaries}

This section establishes the fundamental concepts that underpin CLIP-powered domain generalization (DG) and domain adaptation (DA).
It first introduces the notations (\S\ref{sec_Preliminaries_Notations}) used throughout the survey and then provides the formal definitions and settings (\S\ref{sec_Preliminaries_Definitions}) that serve as the basis for subsequent methodological analysis.

\subsection{Notations} 
\label{sec_Preliminaries_Notations}

We summarize the key notations in Table~\ref{tab:notations-and-descriptions}, which outlines the symbols and formulas used in our analysis.

\subsection{Definitions} 
\label{sec_Preliminaries_Definitions}

This subsection formalizes the core problem settings and terminologies of DG and DA, providing definitions and categorizations that support the methodological discussions in later sections.
It introduces the basic definitions (\S\ref{sec_Preliminaries_Definitions_Problem-Definition}) of DG and DA, describes their common variants according to source availability (\S\ref{sec_Preliminaries_Definitions_SA-SF}) and label-space relationships (\S\ref{sec_Preliminaries_Definitions_CS-PS-OS-OPS}), and concludes with a brief formulation of the CLIP framework (\S\ref{sec_Preliminaries_Definitions_CLIP}).

\begin{table}[b]
\centering
\caption{\textcolor{revision-color_1}{Several commonly used notations and descriptions in this paper.}}
\rowcolors{2}{white}{mytabcolor_4}  
\label{tab:notations-and-descriptions}
\renewcommand{\arraystretch}{1.6}
\resizebox{0.99\linewidth}{!}{
\begin{tabular}{cl}
    \toprule
    
    \textbf{Notation} & \textbf{\makecell[c]{Description}} \\
    \midrule
    
    $f: \mathbb{R}^{H \times W} \mapsto \mathbb{R}^{C}$ 
        & Mapping function from image space to class space. \\
    
    $\mathcal{D}_s = \{D_1, D_2, \dots, D_S\}$ 
        & Set of multiple source domains. \\
    
    $\mathcal{D}_t = \{D_1, D_2, \dots, D_T\}$   
        & Set of unseen target domains. \\
    
    $D_K = \{(x_i, y_i)\}_{i=1}^{N}$ 
        & Dataset of domain $K$ with $N$ samples, where $x_i$ is the input and $y_i$ is the label. \\
    
    $\mathcal{X}_K \in \mathbb{R}^{H \times W}$ 
        & Input image space of domain $K$. \\
    
    $\mathcal{Y}_s \in \mathbb{R}^{C_s}$, $\mathcal{Y}_t \in \mathbb{R}^{C_t}$ 
        & Label spaces of the source and target domains. \\
    
    $C$ 
        & Number of classes in the label space. \\
    \bottomrule
\end{tabular}
}
\end{table}

\begin{figure}[t]
    \centering
    \includegraphics[width=0.99\linewidth]{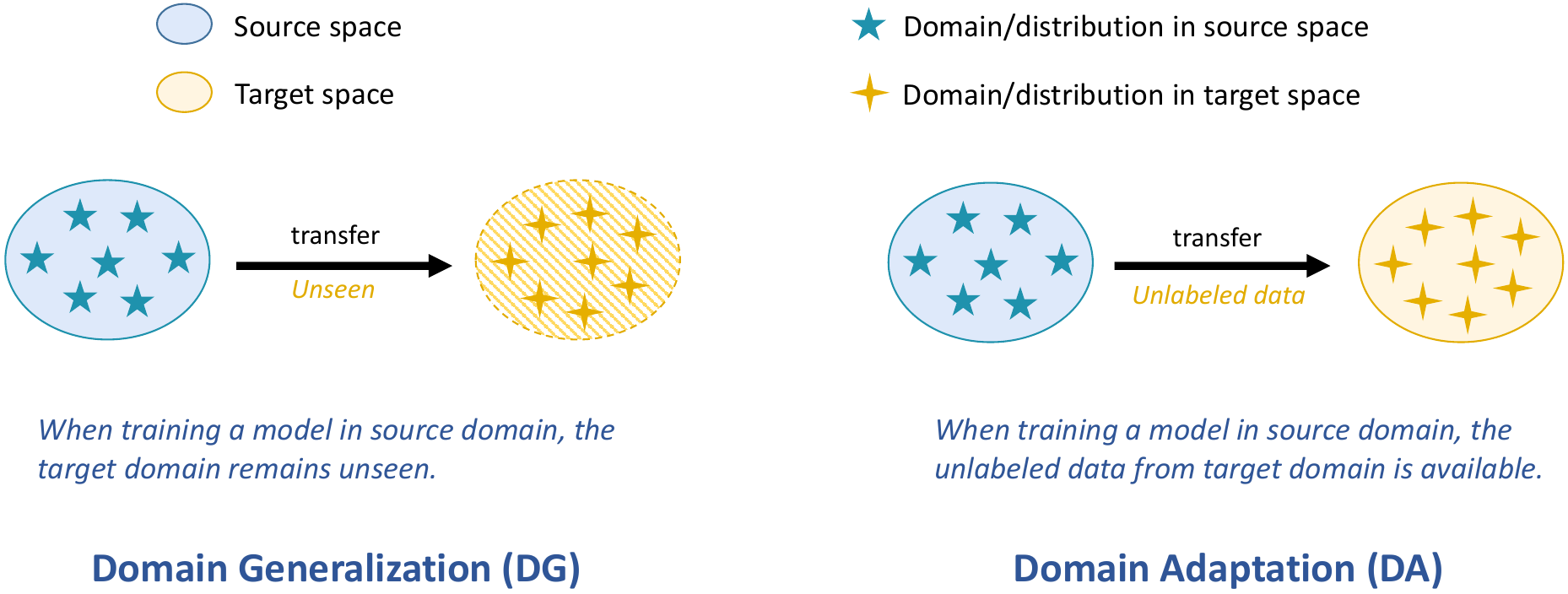}
    \caption{\textcolor{revision-color_1}{Comparison of domain generalization (DG) and adaptation (DA).}}
    \label{fig:Comparision-of-DG-and-DA.}
\end{figure}

\subsubsection{Problem Definition}
\label{sec_Preliminaries_Definitions_Problem-Definition}

Domain Generalization (DG) and Domain Adaptation (DA) both aim to handle distribution shifts between domains. As shown in Fig.~\ref{fig:Comparision-of-DG-and-DA.}, DG trains a model on one or multiple source domains and evaluates it on unseen target domains without any target access. DA, in contrast, trains on a labeled source domain and adapts the model to a target domain, which is typically unlabeled. Thus, DG emphasizes robustness to unseen domains, whereas DA focuses on transferring knowledge to a specific target domain.

\vspace{0.5em}
\noindent\textbf{Definition 1.}
\textit{Domain Generalization (DG). 
Domain generalization refers to methods that train a model $f$ on one or multiple 
labeled source domains $\mathcal{D}_s = \{D_1, D_2, \cdots, D_S\}$. 
The goal is to ensure that the model performs reliably on unseen target domains $\mathcal{D}_t = \{D'_1, D'_2, \cdots, D'_T\}$\textbf{ without accessing any target samples during training}. This setting emphasizes robustness to distribution shifts, enabling the model to operate effectively under diverse and previously unobserved conditions rather than merely fitting the specific datasets used during training.
}
\vspace{0.5em}

\noindent\textbf{Definition 2.}
\textit{Domain Adaptation (DA). 
Domain adaptation involves training a model $f$ on one or multiple labeled source domains $\mathcal{D}_s=\{D_1,D_2,\dots,D_S\}$ and adapting it to a target domain $\mathcal{D}_t=\{x_j\}_{j=1}^{N_t}$ whose distribution differs from that of the source domains. The target domain typically lacks labels or provides only limited supervision, depending on the specific adaptation scenario. The objective is to adjust the knowledge learned from $\mathcal{D}_s$ so that the model remains effective on $\mathcal{D}_t$ despite the presence of distribution shifts. 
Although DA includes multiple supervision levels, this survey focuses on the widely studied \textbf{unsupervised domain adaptation (UDA)} setting, where the target domain contains no labels.
}

\subsubsection{Different Scenarios about Availability of Source Information}
\label{sec_Preliminaries_Definitions_SA-SF}

Source-Available (SA) scenarios provide labeled data from the source domain to enhance adaptation in the target domain and are further divided into Single-Source (SS) and Multi-Source (MS) methods, as shown in Fig.~\ref{fig:Illustration-of-SA.}. SS focuses on transferring knowledge from one labeled source domain, while MS leverages information from multiple labeled source domains for improved adaptation. As illustrated in Fig.~\ref{fig:SF}, within Source-Free (SF), there are two subcategories: Source-Data-Free (SDF) allows for extracting knowledge from the source model without direct access to labeled source data, facilitating effective adaptation. In contrast, Source-Fully-Free (SFF) prohibits any information from the source domain, requiring the model to rely solely on target data, which significantly increases adaptation challenges. Their specific definitions are as follows:

\begin{figure}[t]
    \centering
    \includegraphics[width=0.99\linewidth]{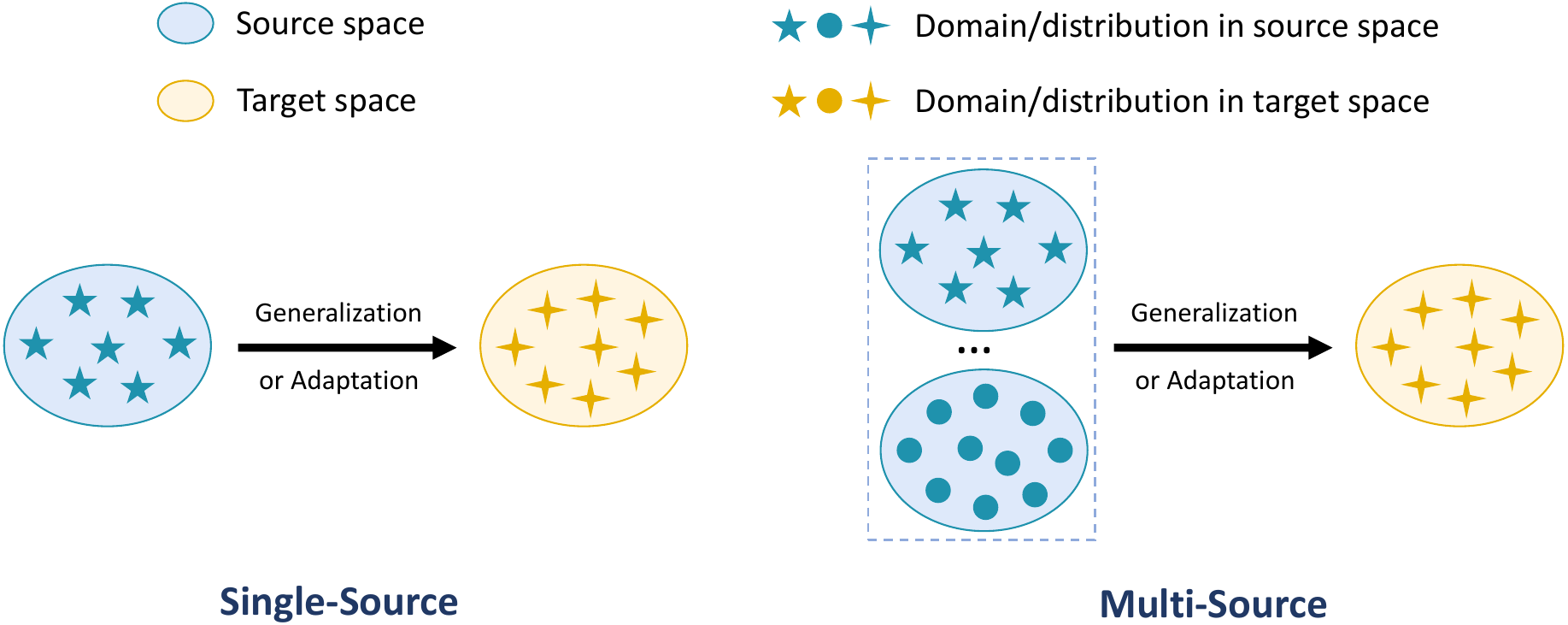}
    \caption{\textcolor{revision-color_1}{Comparison of single-source (SS) and multi-source (MS) scenario.}}
    \label{fig:Illustration-of-SA.}
\end{figure}

\vspace{0.5em}
\noindent\textbf{Definition 3.}
\textit{Source-Available (SA). 
Source-available refers to scenarios in which labeled data from the source domain $\mathcal{D}_s$ is accessible for adaptation. This availability facilitates the transfer of knowledge gained from the source domain to improve performance in the target domain $\mathcal{D}_t$.
}
\vspace{0.5em}

\noindent\textbf{Definition 3.1.}
\textit{Single-Source (SS). 
Single-source-based methods are specifically designed to tackle shifts in data distribution by enabling the transfer of knowledge from a labeled source domain $\mathcal{D}_s$ to the target domain $\mathcal{D}_t$.
}
\vspace{0.5em}

\noindent\textbf{Definition 3.2.}
\textit{Multi-Source (MS). 
Multi-source-based methods employ strategies to address shifts in data distribution by facilitating the transfer of knowledge from multiple labeled source domains $\mathcal{D}_s = \{D_1, D_2, \cdots, D_S\}$ to the target domain $\mathcal{D}_t$.
}
\vspace{0.5em}

\noindent\textbf{Definition 4.}
\textit{Source-Data-Free (SDF). 
Source-data-free refers to a setting in which no data from the source domain $\mathcal{D}_s$ are accessible during adaptation. Instead, only a model pre-trained on the source domain is available. The adaptation process must therefore be carried out using only the unlabeled target domain data $\mathcal{D}_t$, which necessitates effective knowledge transfer without direct access to the source data. 
}
\vspace{0.5em}

\noindent\textbf{Definition 5.}
\textit{Source-Fully-Free (SFF). 
Source-fully-free denotes a strict setting where no information from the source domain $\mathcal{D}_s$ is available. Models rely solely on a general-purpose pretrained foundation model (e.g., CLIP) and perform adaptation directly on the unlabeled target domain $\mathcal{D}_t$, without any supervision or auxiliary knowledge from $\mathcal{D}_s$.
}

\begin{figure}[t]
    \centering
    \includegraphics[width=0.99\linewidth]{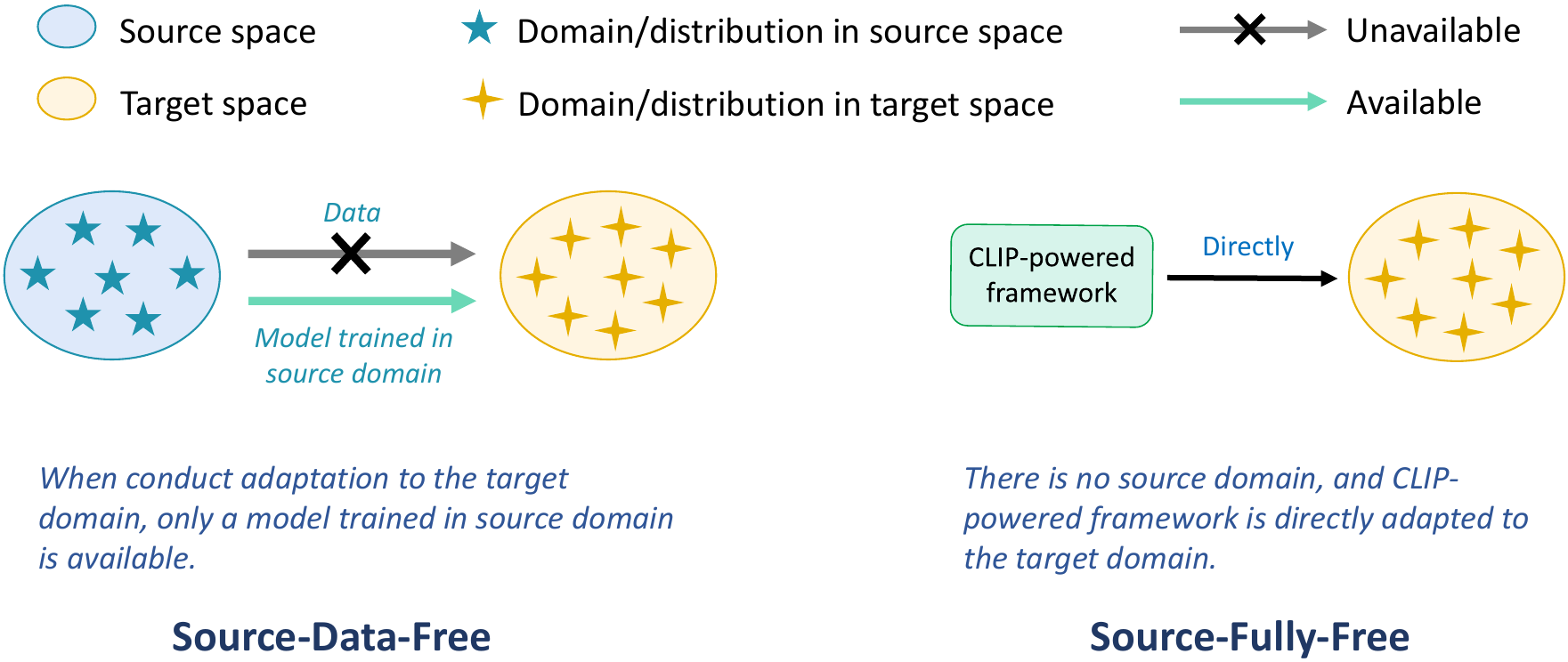}
    \caption{\textcolor{revision-color_1}{Comparison of source-data-free (SDF) and source-fully-free (SFF). (Take single-source scenario as an example.)}}
    \label{fig:SF}
\end{figure}

\subsubsection{Different Scenarios about Connection between the Source Domain and Target Domain}
\label{sec_Preliminaries_Definitions_CS-PS-OS-OPS}

\begin{figure}[t]
    \centering
    \includegraphics[width=0.99\linewidth]{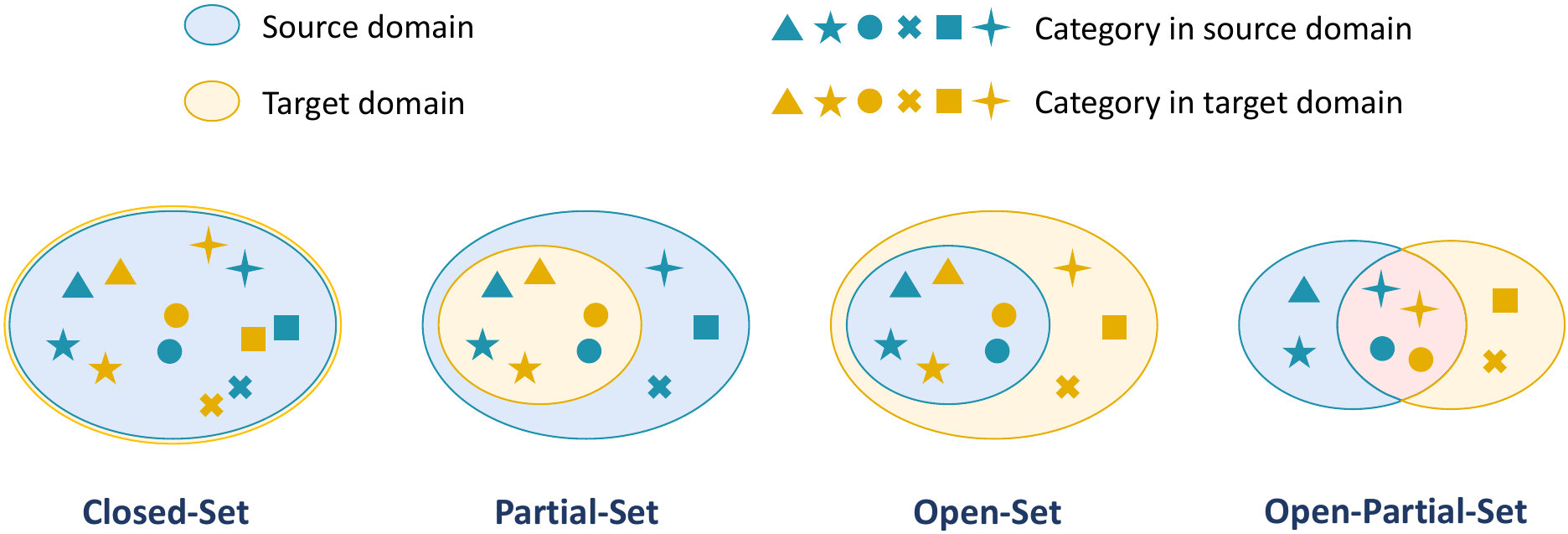}
    \caption{\textcolor{revision-color_1}{Comparison of closed-set (CS), partial-set (PS), open-set scenarios (OS) and open-partial-set (OPS).}}
    \label{fig:Closed-Partial-Open-set}
\end{figure}

As shown in Fig.~\ref{fig:Closed-Partial-Open-set}, there are 4 relationship types that describe the connection between the source domain and target domain: closed-set (CS), partial-set (PS), open-set (OS), and open-partial-set (OPS) scenarios. 
\textbf{\textit{(i)}} In the closed-set (CS) scenario, the source and target domains share an identical set of categories, and the model only needs to focus on minimizing the distribution discrepancy between the two domains. 
\textbf{\textit{(ii)}} In the partial-set (PS) scenario, the target domain is a subset of the source domain's categories,
requiring the model not only to align the shared classes but also to prevent target samples from being misclassified into source-private classes. 
\textbf{\textit{(iii)}} In the open-set (OS) scenario, the target domain includes novel categories besides those seen in the source domain, and the model must identify these unknown classes and group them accordingly. 
\textbf{\textit{(iv)}} In the open-partial-set (OPS) scenario, both domains share some categories while also having their own unique ones, posing a more complex challenge where the model must align shared categories and isolate private ones to handle severe category shift.
Their specific definitions are as follows:

\vspace{0.5em}
\noindent\textbf{Definition 6.}
\textit{Closed-Set (CS). 
Closed-set scenarios refer to situations where the label space of the target domain $ \mathcal{Y}_t $ is the same as that of the source domain $ \mathcal{Y}_s $, such that $ \mathcal{Y}_t = \mathcal{Y}_s $. 
}
\vspace{0.5em}

\noindent\textbf{Definition 7.}
\textit{Partial-Set (PS). 
Partial-set scenarios involve situations where the label space of the target domain $ \mathcal{Y}_t $ is a subset of classes from the source domain $ \mathcal{Y}_s $, 
such that $ \mathcal{Y}_t \subset \mathcal{Y}_s $ and $ \mathcal{Y}_s \setminus \mathcal{Y}_t \neq \emptyset $. 
}

\vspace{0.5em}

\noindent\textbf{Definition 8.} 
\textit{Open-Set (OS). 
Open-set scenarios encompass cases where the label space of the target domain $ \mathcal{Y}_t $ strictly contains the classes from the source domain $ \mathcal{Y}_s $, and introduces additional novel classes, such that $ \mathcal{Y}_s \subset \mathcal{Y}_t$ and $ \mathcal{Y}_t \setminus \mathcal{Y}_s \neq \emptyset $. 
}
\vspace{0.5em}

\noindent\textbf{Definition 9.}
\textit{Open-Partial-Set (OPS). 
Open-partial-set scenarios refer to situations where the label spaces $ \mathcal{Y}_s $, $ \mathcal{Y}_t $ of the source and target domains partially overlap while also having their own private classes, such that $ \mathcal{Y}_s \cap \mathcal{Y}_t \neq \emptyset $, $ \mathcal{Y}_t \nsubseteq \mathcal{Y}_s $, and $ \mathcal{Y}_s \nsubseteq \mathcal{Y}_t $. This indicates that there are some classes present in $ \mathcal{Y}_t $ that are not in $ \mathcal{Y}_s$, while also having classes in $ \mathcal{Y}_s $ that don't appear in $ \mathcal{Y}_t $. 
}
\vspace{0.5em}

\subsubsection{Contrastive Language-Image Pretraining (CLIP)}
\label{sec_Preliminaries_Definitions_CLIP}

\begin{figure}
    \centering
    \subfigure[Training process]{
        \includegraphics[width=0.4\linewidth]{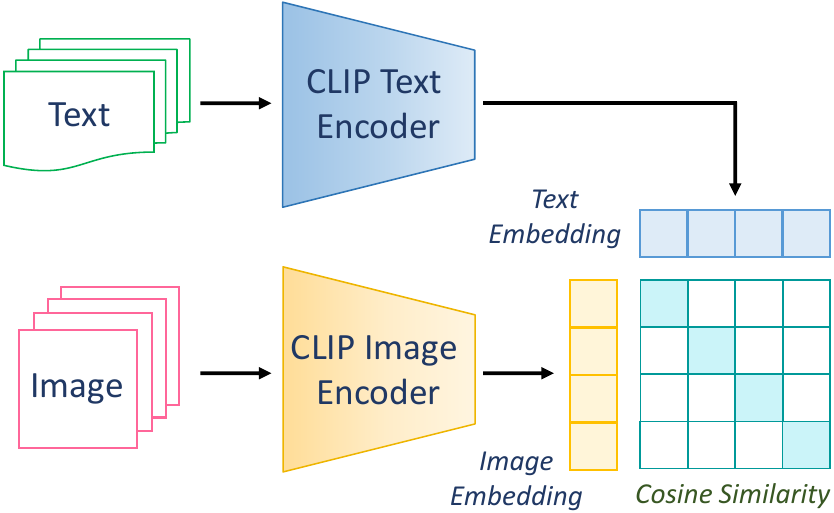}
        \label{fig:CLIP-training}
    }
    \subfigure[Zero-shot ability]{
        \includegraphics[width=0.5\linewidth]{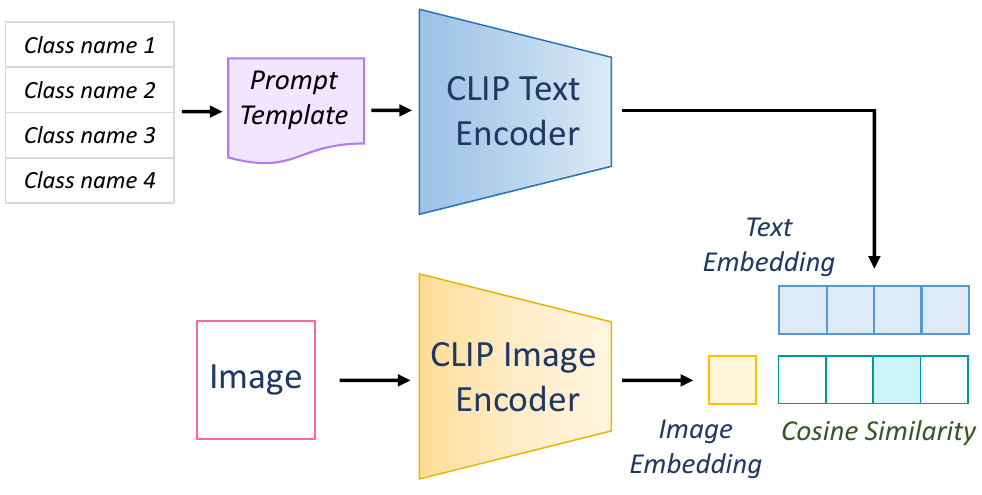}
        \label{fig:CLIP-zero-shot}
    }
\caption{\textcolor{revision-color_1}{The illustration of (a). training process of CLIP (taking 4 class names and 4 images as example), and (b). zero-shot ability of CLIP (taking 4 class names as example) \cite{2021_ICML_CLIP}.}}
\label{fig:CLIP-traing-and-zero-shot}
\end{figure}

CLIP learns visual concepts by jointly training on images and their textual descriptions using contrastive learning, aligning images and text in a shared space, as shown in Fig.~\ref{fig:CLIP-training}. This approach provides flexibility and strong generalization, particularly in zero-shot learning and domain adaptation tasks.
The model is based on a transformer architecture and utilizes a contrastive loss function to align image and text embeddings, enabling it to perform well across different tasks and domains without requiring task-specific fine-tuning, with its zero-shot capability (as shown in Fig.~\ref{fig:CLIP-zero-shot}) allowing it to generalize to new tasks and domains directly from pre-trained models~\cite{2021_ICML_CLIP}.
\vspace{0.5em}

\vspace{0.5em}
\noindent\textbf{Definition 10.}  
\textit{CLIP (Contrastive Language-Image Pretraining).}  
Let $\mathcal{I} = \{I_1, I_2, \dots, I_N\}$ represent a set of images, and $\mathcal{T} = \{T_1, T_2, \dots, T_N\}$ represent a corresponding set of textual descriptions. CLIP learns a joint embedding space $\mathbb{R}^d$ by projecting each image and its associated text into this space. The objective is to maximize the similarity of paired image--text embeddings while minimizing that of mismatched pairs. This is achieved using the following contrastive loss ~\cite{2019_ICLR_DIM, 2020_ICML_SimCLR, 2020_CVPR_MoCo, 2022_PMLR_ConVIRT}:
\textcolor{revision-color_1}{
\begin{equation}
\mathcal{L}_{\text{image}\rightarrow\text{text}}
= -\frac{1}{N} \sum_{i=1}^{N}
\log
\frac{\exp(\mathbf{v}_i^{\top}\mathbf{t}_i / \tau)}
{\sum_{j=1}^{N} \exp(\mathbf{v}_i^{\top}\mathbf{t}_j / \tau)} ,
\end{equation}
\begin{equation}
\mathcal{L}_{\text{text}\rightarrow\text{image}}
= -\frac{1}{N} \sum_{i=1}^{N}
\log
\frac{\exp(\mathbf{t}_i^{\top}\mathbf{v}_i / \tau)}
{\sum_{j=1}^{N} \exp(\mathbf{t}_i^{\top}\mathbf{v}_j / \tau)} ,
\end{equation}
\begin{equation}
\mathcal{L}_{c}
= \frac{1}{2}
\left(
\mathcal{L}_{\text{image}\rightarrow\text{text}}
+
\mathcal{L}_{\text{text}\rightarrow\text{image}}
\right) ,
\end{equation}
}
where $\mathbf{v}_i$ and $\mathbf{t}_i$ denote the image and text embeddings of the $i$-th pair in $\mathbb{R}^d$, and $\tau$ is a temperature parameter. This symmetric loss encourages paired samples to be close in the embedding space while pushing apart non-matching pairs.

\section{Domain Generalization (DG)}
\label{sec_DG}

Traditional (non-CLIP based) methods have been proposed to achieve domain generalization (DG) from the perspective of domain-invariant~\cite{2016_ECCV_Deep-CORAL, 2018_arXiv_Generalizing-Across-Domains-via-Cross-Gradient-Training, 2022_CVPR_Causality-Inspired-Representation-Learning-for-Domain-Generalization, 2023_CVPR_DomainDrop, 2023_CVPR_Modality-Agnostic-Debiasing-for-Single-Domain-Generalization, 2023_CVPR_Decompose-Adjust-Compose}, data augmentation~\cite{2020_ECCV_Digits-DG-dataset_Learning-to-Generate-Novel-Domains-for-Domain-Generalization, 2021_CVPR_A-Fourier-based-Framework-for-Domain-Generalization, 2021_arXiv_Domain-Generalization-with-Mixstyle, 2022_CVPR_Style-Neophile, 2023_CVPR_Progressive-Random-Convolutions-for-Single-Domain-Generalization}, learning strategies~\cite{2018_CVPR_Domain-Generalization-with-Adversarial-Feature-Learning, 2020_ECCV_Self-Challenging-Improves-Cross-Domain-Generalization, 2021_NeurIPS_SWAD, 2023_CVPR_Sharpness-Aware-Gradient-Matching-for-Domain-Generalization, 2023_ICCV_Flatness-Aware-Minimization-for-Domain-Generalization, 2023_TCSVT_Instance-Paradigm-Contrastive-Learning-for-Domain-Generalization}, and etc.

CLIP-based DG seeks to ensure robust performance in unseen target domains by leveraging CLIP’s transferable vision–language representations.
This section reviews recent advances from two key perspectives: prompt optimization (\S\ref{sec_DG_prompt-optimization}) and CLIP as backbone or encoder (\S\ref{sec_DG_CLIP-as-backbone-or-encoder}), followed by comparative analyses and discussions of emerging trends.

\subsection{Prompt Optimization}
\label{sec_DG_prompt-optimization}

\begin{figure}[t!]
    \centering
    \includegraphics[width=0.72\linewidth]{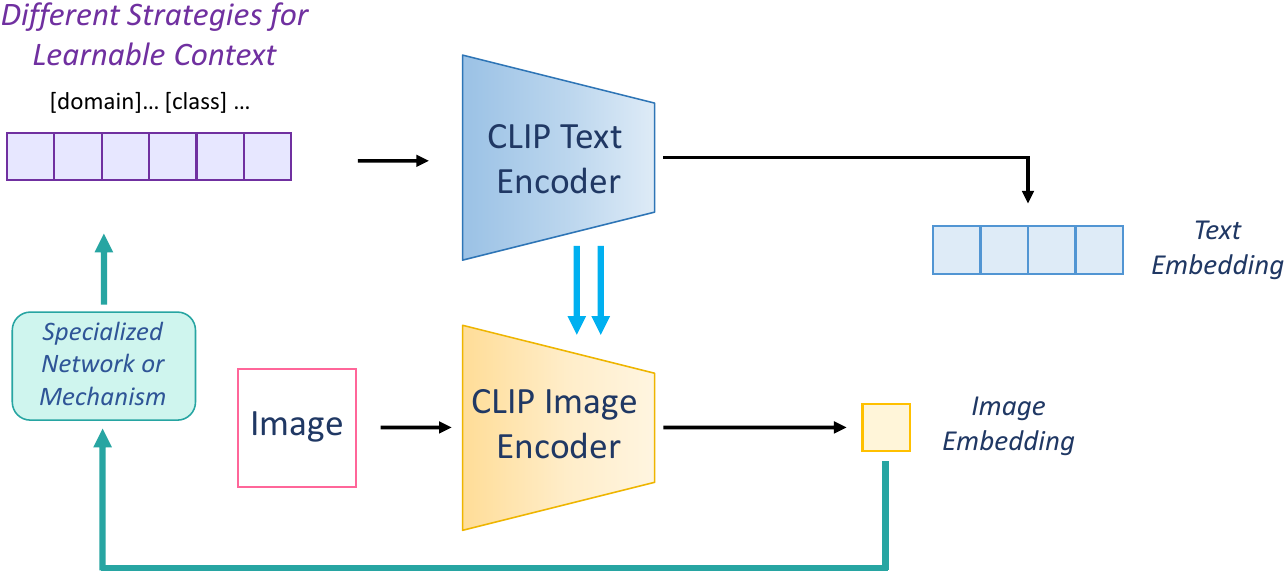}
    \caption{\textcolor{revision-color_1}{The different ways about prompt learning optimization \cite{2022_IJCV_CoOp, 2022_CVPR_CoCoOp, 2023_CVPR_MaPLe, 2024_AAAI_LAMM, 2024_CVPR_ODG-CLIP}.}}
    \label{fig:Prompt-Learning-Optimization}
\end{figure}

Prompt optimization methods \cite{2022_IJCV_CoOp, 2022_CVPR_CoCoOp, 2023_CVPR_KgCoOp, 2023_ICCV_ProGrad, 2023_CVPR_MaPLe, 2024_AAAI_LAMM} focus on refining prompts in CLIP to improve task performance, as illustrated in Fig.~\ref{fig:Prompt-Learning-Optimization}. Recent approaches automate prompt tuning and enhance context representation, enabling more effective utilization of pre-trained models across diverse applications.

CoOp~\cite{2022_IJCV_CoOp} introduces context optimization, which replaces manually designed prompts with continuous learnable context vectors while keeping the pre-trained parameters frozen. Building on prior work in prompting techniques~\cite{2020_arXiv_AutoPrompt, 2020_ACL_How-Can-We-Know-What-Language-Models-Know, 2021_arXiv_Factual-Probing-is-Mask-Learning-vs-Learning-to-Reacll, 2021_ICML_ALIGN, 2022_NeurIPS_CLOOB, 2021_arXiv_Prefix-Tuning, 2022_CVPR_FLAVA, 2021_arXiv_Florence}, it learns two forms of context embeddings: a unified context shared across all categories and a class-specific context (CSC) customized for each class, enabling CLIP to capture fine-grained semantics without manual prompt engineering.
Inspired by prompt learning in NLP~\cite{2020_arXiv_Making-Pre-trained-Language-Models-Better-Few-Shot-Learners, 2021_arXiv_The-Power-of-Scale-for-Parameter-Efficient-Prompt-Tuning} and CV~\cite{2022_ECCV_Prompting-Visual-Language-Models-for-Efficient-Video-Understanding, 2022_DenseCLIP, 2021_arXiv_CPT, 2022_CVPR_PointCLIP}, \citet{2022_CVPR_CoCoOp} further develops Conditional Context Optimization (CoCoOp), using a lightweight neural network to generate input-conditional tokens for each image, enhancing parameter efficiency.
Subsequently,
KgCoOp~\cite{2023_CVPR_KgCoOp} introduces Knowledge-guided context optimization to enhance the generalization of learnable prompts for unseen classes. Building on the widespread use of prompt tuning to adapt pre-trained models~\cite{2019_arXiv_Language-Models-as-Knowledge-Bases, 2021_NeurIPS_Multimodal-Few-shot-Learning-with-Frozen-Language-Models, 2022_FTCGV_Vision-language-Pre-training--Basics-Recent-Advances-and-Future-Trends, 2022_ECCV_Visual-Prompt-Tuning}, it regularizes learnable prompts toward hand-crafted templates via knowledge and contrastive constraints, preserving semantics and improving generalization.
In parallel,
ProGrad~\cite{2023_ICCV_ProGrad} addresses overfitting by selectively updating prompts whose gradients align with general knowledge in VLMs, thereby reducing overfitting and preserving essential semantics to improve prompt tuning performance, outperforming existing approaches~\cite{2020_arXiv_Making-Pre-trained-Language-Models-Better-Few-Shot-Learners, 2021_arXiv_LFPT5, 2022_IJCV_CoOp, 2022_CVPR_CoCoOp}.
Meanwhile,
MaPLe~\cite{2023_CVPR_MaPLe} is the first multimodal prompting framework for fine-tuning CLIP. It aligns vision and language modalities by coupling image and text prompts via a conditioning function, enabling joint optimization through cross-modal gradient propagation.
Most Recently,
LAMM~\cite{2024_AAAI_LAMM} uses trainable category tokens and optimizes $\langle CLASS \rangle$ embeddings through gradient-based search. A hierarchical loss aligns representations across multiple spaces, improving generalization while maintaining CLIP’s semantic consistency.

\subsection{CLIP is Adopted as Backbone or Encoder}
\label{sec_DG_CLIP-as-backbone-or-encoder}

\begin{figure}[t!]
    \centering
    \includegraphics[width=0.99\linewidth]{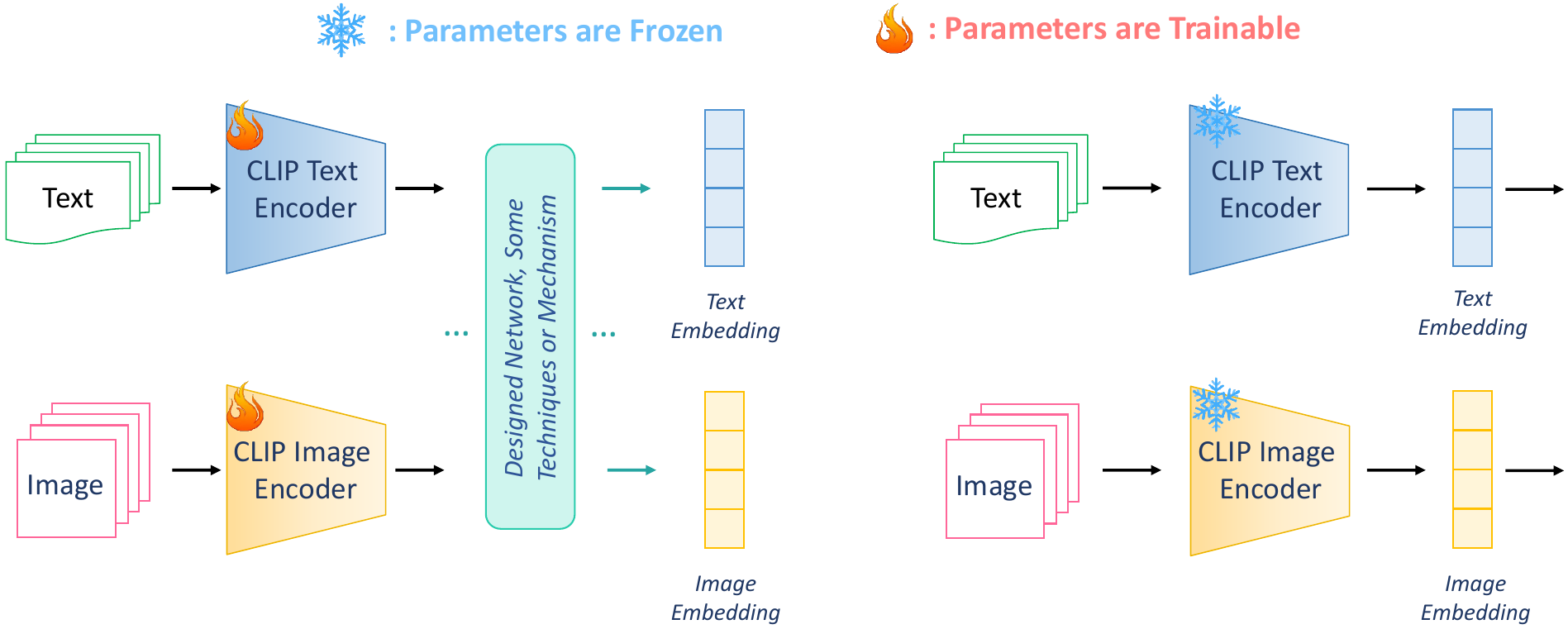}
    \caption{\textcolor{revision-color_1}{The illustration of (Left) CLIP as the backbone and the text encoder or image encoder is trainable, (Right) CLIP as the encoder and the parameters of text encoder and image encoder are frozen.}}
    \label{fig:CLIP_as_backbone_or_encoder}
\end{figure}

CLIP is commonly integrated as a backbone or encoder in domain generalization approaches, serving key roles in feature extraction and performance improvement. There are two primary adoption methods for CLIP, as illustrated in Fig.~\ref{fig:CLIP_as_backbone_or_encoder}.
\textbf{\textit{The first method}} (shown on the left of Fig.~\ref{fig:CLIP_as_backbone_or_encoder}) involves training both CLIP Image and Text Encoders, often combined with task-specific architectures. This allows the model to adapt to downstream tasks and capture domain-specific patterns, enhancing generalization.
\textbf{\textit{The second method}} (shown on the right of Fig.~\ref{fig:CLIP_as_backbone_or_encoder}) freezes the CLIP encoders, using them directly for embedding extraction. This approach leverages CLIP’s pre-trained knowledge, offering computational efficiency and preventing overfitting, especially with limited training resources.

In both methods, CLIP serves as a flexible feature extractor, whether as a dynamic or static model, providing robust and semantically rich embeddings that facilitate effective cross-domain learning.
In this survey, we primarily examine CLIP-based domain generalization from the perspective of source availability, encompassing both source-available (SA) (\S\ref{sec_DG_CLIP-as-backbone-or-encoder_SA}) and source-free (SF) (\S\ref{sec_DG_CLIP-as-backbone-or-encoder_SF}) scenarios.

\subsubsection{Source-Available (SA)}
\label{sec_DG_CLIP-as-backbone-or-encoder_SA}

This subsection further organizes source-available (SA) methods into \textcolor{revision-color_1}{single-source (\S\ref{sec_DG_CLIP-as-backbone-or-encoder_SA_SS}) and multi-source (\S\ref{sec_DG_CLIP-as-backbone-or-encoder_SA_MS_CSDG}, \S\ref{sec_DG_CLIP-as-backbone-or-encoder_SA_MS_OSDG}) categories,} highlighting their methodological characteristics and representative designs.

\paragraph{\textbf{Single-Source Closed-Set Domain Generalization (SS-CSDG)}}
\label{sec_DG_CLIP-as-backbone-or-encoder_SA_SS}

\textit{In SS-CSDG, the model $ f: \mathbb{R}^{H \times W} \mapsto \mathbb{R}^{C} $ is trained on labeled data from a single source domain $ \mathcal{D}_s = \{(x_i, y_i)\}_{i=1}^{N_s} $. The objective is to learn domain-invariant representation on the source domain while maximizing generalization to unseen target domain $ \mathcal{D}_t = \{x_j\}_{j=1}^{N_t} $, where the label space of the target domain $ \mathcal{Y}_t $ is identical to that of the source domain $ \mathcal{Y}_s $, i.e., $ \mathcal{Y}_t = \mathcal{Y}_s $.}

Existing studies enhance single-source generalization by leveraging limited domain knowledge to learn invariant representations and improve robustness to unseen target domains.
They can be broadly grouped into two categories: \textit{(i) prompt-driven optimization methods} that refine or generate prompts to enhance domain robustness \cite{2023_ICCV_DAPT, 2023_ICCV_PromptSRC, 2024_arXiv_SPG, 2024_arXiv_LDFS, 2024_arXiv_GalLoP, 2025_CVPR_FrogDogNet_FrogDogNet=Fourier-Frequency-Retained-Visual-Prompt-Output-Guidance-for-Domain-Generalization-of-CLIP-in-Remote-Sensing}, and \textit{(ii) architecture-enhanced alignment methods} that extend CLIP’s backbone with adapters or feature fusion modules \cite{2023_ICCV_BorLan, 2024_CVPR_MMA, 2024_WACV_StyLIP}.

\textbf{\textit{(i) Prompt-driven optimization methods.}}
DAPT~\cite{2023_ICCV_DAPT} optimizes both textual and visual prompts to match the data distribution. Unlike traditional methods~\cite{2022_arXiv_Visual-Prompting--Modifying-Pixel-Sapce-to-Adapt-Pre-trained-Models, 2022_ECCV_Visual-Prompt-Tuning, 2022_OpenReview_Prompt-Learning-with-Optimal-Transport-for-Vision-Language-Models, 2022_CVPR_Prompt-Distribution-Learning, 2022_arXiv_Unified-Vision-and-Language-Prompt-Learning, 2024_CVPR_Multitask-Vision-Language-Prompt-Tuning}, it uses inter- and intra-dispersion losses to guide the learning of diverse text embeddings and compact image embeddings within each class.
Similarly,
PromptSRC~\cite{2023_ICCV_PromptSRC} is a self-regularization framework for prompt learning inspired by network regularization techniques
~\cite{2014_JMLR_DropOut, 2015_arXiv_Batch-Normalization, 2017_arXiv_MixUp, 2016_CVPR_Rethinking-the-Inception-Architecture-for-Computer-Vision, 2017_arXiv_Decoupled-Weight-Decay-Regularization, 2019_ICCV_CutMix, 2020_CVPRW_RandAugment, 2021_ICML_Improved-ODD-Generalization-via-Adversarial-Training-and-Pretraining, 2022_ECCV_Cross-Domain-Ensemble-Distillation-for-Domain-Generalization, 2022_NeurIPS_Patching-Open-vocabulary-Models-by-Interpolating-Weights, 2022_CVPR_Robust-Fine-tuning-of-Zero-shot-Models}. 
It regularizes prompt learning by aligning with the frozen model and enforcing intra-prompt consistency to improve stability and cross-modal generalization.
In parallel,
SPG~\cite{2024_arXiv_SPG} reframes prompt learning from a generative view. It trains with domain-specific soft prompt labels and uses a generative model to produce instance-specific prompts for unseen domains during inference.
Meanwhile,
LDFS~\cite{2024_arXiv_LDFS} generates diverse domain features via text-guided instance-conditional augmentation. It incorporates a pairwise regularizer for feature coherence and uses stochastic text augmentation to bridge the modality gap~\cite{2021_ICCV_StyleCLIP, 2022_TOG_StyleGAN-NADA, 2022_CVPR_CLIPStyler, 2022_CVPR_DiffusionCLIP}.
In contrast,
GalLoP~\cite{2024_arXiv_GalLoP} integrates global and local visual features to create diverse prompts. Local prompts align with sparse image regions, enabling precise text-to-image matching for fine-grained semantics. It refines textual alignment of local visual features
~\cite{2022_arXiv_PLOT, 2022_NeurIPS_DualCoOp, 2022_ECCV_Extract-Free-Dense-Labels-from-CLIP, 2023_arXiv_GL-MCM_Zero-shot-In-disstribution-Detection-in-Multi-object-Settings-Using-Vision-language-Foundation-Models, 2023_CVPR_MaskCLIP, 2023_NeurIPS_LoCoOp} 
using linear projection for few-shot learning. To enhance diversity, it combines global and localized prompts with a "prompt dropout" strategy~\cite{2016_ICML_Dropout-as-A-Bayesian-Approximation--Representing-Model-Uncertainty-in-Deep-Learning, 2014_JMLR_DropOut}, and uses a multiscale approach to capture broader semantic details.
Recently,
FrogDogNet~\cite{2025_CVPR_FrogDogNet_FrogDogNet=Fourier-Frequency-Retained-Visual-Prompt-Output-Guidance-for-Domain-Generalization-of-CLIP-in-Remote-Sensing} incorporates a fourier filter block to retain low-frequency invariant components and suppress background noise, combined with a remote sensing prompt alignment loss to enhance cross-domain feature stability.

\textbf{\textit{(ii) Architecture-enhanced alignment methods.}}
BorLan~\cite{2023_ICCV_BorLan} uses pre-trained language models to align visual and linguistic features, helping the vision model learn semantic-aware, domain-agnostic representations. It is model-agnostic, supporting various visual and language backbones.
In parallel,
MMA~\cite{2024_CVPR_MMA} introduces the multimodal adapter for VLMs to improve alignment between text and vision representations, inspired by previous work on adapter methods
~\cite{2019_ICML_Parameter-efficient-Transfer-Learning-for-NLP, 2019_ICML_BERT-and-PALs, 2020_NeurIPS_TinyTL, 2021_arXiv_LoRA, 2022_NeurIPS_AdaptFormer, 2022_arXiv_Vision-Transformer-Adapter-for-Dense-Predictions, 2024_IJCV_CLIP-Adapter, 2024_AAAI_T2i-Adapter}. It aggregates features from both branches into a shared space for gradient communication, applying only to higher layers to balance discrimination and generalization.
In contrast,
StyLIP~\cite{2024_WACV_StyLIP} leverages CLIP’s frozen vision encoder and lightweight projectors to extract multiscale style features, enhancing hierarchical domain knowledge and generalization.

\paragraph{\textbf{Multi-Source Closed-Set Domain Generalization (MS-CSDG)}}
\label{sec_DG_CLIP-as-backbone-or-encoder_SA_MS_CSDG}

\textit{In MS-CSDG, the model $ f: \mathbb{R}^{H \times W} \mapsto \mathbb{R}^{C} $ is trained on labeled data from multiple source domains $ \mathcal{D}_s = \{D_1, D_2, \cdots, D_S\} $. The objective is to learn domain-invariant representations across the source domains while maximizing generalization to unseen target domain $ \mathcal{D}_t = \{x_j\}_{j=1}^{N_t} $, where the label space of the target domain $ \mathcal{Y}_t $ is identical to that of the source domains $ \mathcal{Y}_s $, i.e., $ \mathcal{Y}_t = \mathcal{Y}_s $. 
}

Existing studies improve multi-source generalization by leveraging knowledge from multiple domains to learn domain-invariant and semantically consistent representations.
They can be grouped into three types: 
\textit{(i) distillation-based methods} that transfer knowledge from vision–language teachers or auxiliary branches \cite{2023_ICCV_RISE, 2024_CVPR_VL2V-ADiP, 2024_Access_CAL}; 
\textit{(ii) prompt-driven methods} that refine or generate domain-aware prompts to improve adaptability \cite{2023_TJSAI_DPL, 2024_arXiv_SPG, 2024_CVPR_Any-Shift-Prompting, 2024_CVPR_DPR, 2025_arXiv_PADG_Prompt-Disentanglement-via-Language-Guidance-and-Representation-Alignment-for-Domain-Generalization}; 
and \textit{(iii) architecture-enhanced methods} that employ adapters, model merging, or feature fusion to strengthen cross-domain consistency \cite{2024_WACV_StyLIP, 2024_CVPR_ODG-CLIP, 2023_ICML_CLIPood, 2025_arXiv_HAM_Harmonizing-and-Merging-Source-Models-for-CLIP-based-Domain-Generalization, 2025_TCSVT_ClipMix_ClipMix-for-Domain-Generalization, 2025_WACV_MoA_Domain-Generalization-using-Large-Pretrained-Models-with-Mixture-of-Adapters}.

\textbf{\textit{(i) Distillation-based methods.}}
RISE~\cite{2023_ICCV_RISE} distills knowledge from a vision-language teacher by aligning student image features with teacher text features using absolute and relative distance losses for domain generalization.
Similarly,
\citet{2024_CVPR_VL2V-ADiP} proposes VL2V-SD for text-guided self-distillation and VL2V-ADiP for black-box feature fusion, both enhancing OOD generalization.
Meanwhile,
CAL~\cite{2024_Access_CAL} enhances CLIP-based domain generalization by employing CAFT during training and ETTA during inference, thereby improving model robustness across domains.

\textbf{\textit{(ii) Prompt-driven methods.}}
DPL~\citep{2023_TJSAI_DPL} employs a lightweight three-layer MLP–based prompt generator to enhance accuracy in domain generalization while keeping the parameter size compact.
Similarly,
\citet{2024_arXiv_SPG} applies to this scenario by generating soft prompts from multiple source domains, enhancing the model’s generalization to unseen domains.
Meanwhile,
Any-Shift Prompting~\cite{2024_CVPR_Any-Shift-Prompting} enhances CLIP’s generalization through hierarchical prompts and a pseudo-shift mechanism, achieving improved adaptability without incurring additional training costs.
Furthermore,
DPR~\cite{2024_CVPR_DPR} proposes a prompt tuning framework that disentangles text and visual prompts, enhancing generalization through domain-specific prototypes and invariant prediction fusion.
To mitigate CLIP’s bias toward source-domain–specific features, \citet{2025_CVPR_Diverse-Text-Prompt_Domain-Generalization-in-CLIP-via-Learning-with-Diverse-Text-Prompts} introduces diverse text prompts that emulate multiple domain contexts, enabling the model to learn domain-invariant representations through suppression, consistency, and diversification.
More recently,
PADG~\cite{2025_arXiv_PADG_Prompt-Disentanglement-via-Language-Guidance-and-Representation-Alignment-for-Domain-Generalization} is an extended version of DPR~\cite{2024_CVPR_DPR} that leverages controllable language prompts to disentangle invariant features and introduces worst explicit representation alignment to enhance visual–textual consistency.

\textbf{\textit{(iii) Architecture-level enhancement methods.}}
StyLIP~\cite{2024_WACV_StyLIP} disentangles visual style and content information from CLIP’s frozen vision encoder and employs lightweight style and content projectors to generate domain-aware prompts
In parallel,
\citet{2024_CVPR_ODG-CLIP} focuses on multi-source open-set domain generalization, treating closed-set DG as a special case. It demonstrates superior performance on MS-CSDG tasks compared to existing benchmarks.
Similarly,
Mixup-CLIPood~\cite{2024_AIEA_Mixup-CLIPood} improves multi-modal domain generalization for object recognition by incorporating mix-up loss and leveraging larger vision–language backbones to achieve stronger performance.
Subsequently,
HAM~\cite{2025_arXiv_HAM_Harmonizing-and-Merging-Source-Models-for-CLIP-based-Domain-Generalization} adopts a learning-then-merging paradigm that trains separate CLIP encoders for each source domain and harmonizes them through conflict-aware model merging.
Moreover,
ClipMix~\cite{2025_TCSVT_ClipMix_ClipMix-for-Domain-Generalization} leverages CLIP as external knowledge to guide adaptive image mixing and label shift, combining pixel- and feature-level enhancements to preserve semantic integrity.
Recently,
MoA~\cite{2025_WACV_MoA_Domain-Generalization-using-Large-Pretrained-Models-with-Mixture-of-Adapters} performs parameter-efficient fine-tuning on CLIP using a mixture-of-adapters architecture, where adapters of different capacities and a learnable router balance regularization strength to handle distribution shifts.

\paragraph{\textbf{Multi-Source Open-Set Domain Generalization (MS-OSDG)}}
\label{sec_DG_CLIP-as-backbone-or-encoder_SA_MS_OSDG}

\textit{In MS-OSDG, the model $ f: \mathbb{R}^{H \times W} \mapsto \mathbb{R}^{C} $ is trained on labeled data from multiple source domains $ \mathcal{D}_s = \{D_1, D_2, \cdots, D_S\} $. The objective is to learn domain-invariant representations across the source domains while generalizing to an unseen target domain $ \mathcal{D}_t = \{x_j\}_{j=1}^{N_t} $, where the label space of the source domains $ \mathcal{Y}_s $ is a subset of the target label space $ \mathcal{Y}_t $, i.e., $ \mathcal{Y}_s \subset \mathcal{Y}_t $.}

The scene of MS-OSDG has relatively little exploration at present, and there are not many traditional methods~\cite{2021_CVPR_Open-Domain-Generalization-with-Domain-augmented-Meta-learning, 2021_ICIP_Open-set-Domain-Generalization-via-Metric-Learning, 2023_ICCV_Generalizaiton-Decision-Boundaries--Dualistic-Meta-learning-for-Open-Set-Domain-Generalization, 2024_IJCNN_Simple-Domain-Generalization-Methods-are-Strong-Baselines-for-Open-Domain-Generalization} either.
Existing approaches in this scenario extend CLIP’s ability to recognize unseen categories while maintaining robustness under domain shifts, mainly improving semantic alignment through diffusion-based synthesis, prompt learning, and knowledge distillation.

CLIPood~\cite{2023_ICML_CLIPood} adapts CLIP to OOD settings via margin metric softmax with class-adaptive margins and a beta-weighted ensemble of zero-shot and fine-tuned models.
Subsequently,
ODG-CLIP~\cite{2024_CVPR_ODG-CLIP} utilizes a stable diffusion model~\cite{2022_CVPR_High-Resolution_Image-Synthesis_with_Latent_Diffusion_Models} with faster inference speed than existing text-to-image generation methods~\cite{2022_MindTrek_The-Creativity-of-Text-to-Image-Generation, 2021_ICML_Zero-Shot-Text-to-Image-Generation, 2022_NeurIPS_Photorealistic-Text-to-Image-Diffusion-Models-with-Deep-Language-Understanding} to generate proxy images for unknown classes and employs a style-aware prompt mechanism to enhance domain discrimination and semantic alignment.
In parallel,
SCI-PD~\cite{2024_CVPR_SCI-PD_PracticalDG=Perturbation-Distillation-on-Vision-Language-Models-for-Hybrid-Domain-Generalization} proposes perturbation distillation to transfer knowledge from VLMs to lightweight vision models, improving robustness via score, class, and instance perspectives. It also introduces the Hybrid Domain Generalization (HDG) benchmark and H2-CV metric for broader evaluation.
More recently,
OSLoPrompt~\cite{2025_CVPR_OSLoPrompt_OSLoPrompt=Bridging-Low-Supervision-Challenges-and-Open-Set-Domain-Generalization-in-CLIP} combines domain-agnostic prompt learning with visually guided semantic attributes and fine-grained pseudo-open sample synthesis using GPT-4o~\cite{2023_arXiv_GPT-4_GPT-4-technical-Report} and stable diffusion.
MetaPrompt~\cite{2025_TMLR_MetaPrompt_Meta-Learning-to-Teach-Semantic-Prompts-for-Open-Domain-Generalization-in-Vision-Language-Models} leverages meta-learning to enhance robustness by designing novel prompts with semantic attention and unsupervised contrastive loss, improving the differentiation between known and unknown classes across multiple domains.

\subsubsection{Source-Free (SF)}
\label{sec_DG_CLIP-as-backbone-or-encoder_SF}

This section further organizes source-free methods into domain generalization (DG) approaches, highlighting their methodological characteristics and summarizing representative works.

\paragraph{\textbf{Source-(Fully)-Free Domain Generalization (S(F)F-DG)}}

\textit{
In S(F)F-DG, the model $ f: \mathbb{R}^{H \times W} \mapsto \mathbb{R}^{C} $ is generalized to unseen target domains $ \mathcal{D}_t = \{x_j\}_{j=1}^{N_t} $ without relying on any information from any source domains $ \mathcal{D}_s $. 
In this context, models must operate solely with the pretrained backbone (e.g., CLIP) together with task-level semantic information, such as the label space or category names of the target dataset, to achieve generalization in the target domain.
}

Existing methods achieve this by leveraging CLIP’s inherent generalization capacity, either constructing domain-invariant prompt representations or introducing style-aware text–image adaptation to enable robust transfer solely from category-level supervision.

DUPRG~\cite{2022_arXiv_DUPRG} learns domain-invariant text representations by aggregating embeddings from a domain bank through a domain-unified prompt generator.
Subsequently,
PromptStyler~\cite{2023_ICCV_PromptStyler} is the first attempt to synthesize a variety of styles in a joint vision-language space via prompts, effectively tackling source-free domain generalization. 
Similarly,
DPStyler~\cite{2024_arXiv_DPStyler} designs style generation and style removal modules to tackle these challenges. The style generation module updates all styles at each training epoch, while the style removal module mitigates variations in the encoder's output features that result from input styles.
In parallel,
PromptTA~\cite{2024_arXiv_PromptTA} integrates a text adapter to address the challenging SF-DG task. This approach includes a style feature resampling module aimed at effectively capturing the distribution of style features, employing resampling techniques to ensure comprehensive coverage across various domains.
Most recently,
BatStyler~\cite{2025_TCSVT_BatStyler_BatStyler=Advancing-Multi-category-Style-Generation-for-Source-free-Domain-Generalization} builds on PromptStyler~\cite{2023_ICCV_PromptStyler}, introducing coarse semantic generation via GPT-4 \cite{2023_arXiv_GPT-4_GPT-4-technical-Report} to relax excessive consistency constraints and uniform style generation based on neural collapse to create uniformly distributed style templates.


\begin{table}[t!]
\centering
\caption{\textcolor{revision-color_1}{Performance (accuracy, (\%)) comparison of representative DG methods, all using CLIP with ViT-B/16 as the vision encoder backbone. SF: source-free. (We report the result from their related original paper.)}}
\renewcommand{\arraystretch}{1.4}
\scalebox{0.72}{
    \begin{tabular}{c|c|cccc|c}
    \toprule
    Method & SF    & PACS  & VLCS  & Office-Home & DomainNet & AVG \\
    \midrule
    
    CLIP \cite{2021_ICML_CLIP} & N/A   & 95.8  & 80.6  & 78.6  & 54.1  & 77.3  \\
    
    CoOp \cite{2022_IJCV_CoOp} & $\times$    & 97.0  & 83.0  & 81.1  & 59.5  & 80.2  \\
    
    CoCoOp \cite{2022_CVPR_CoCoOp} & $\times$    & 96.7  & 83.6  & 80.7  & 59.7  & 80.2  \\
    
    DPL \cite{2023_TJSAI_DPL}  & $\times$    & 97.1  & 84.0  & 83.0  & 59.9  & 81.0  \\
    
    ProGrad \cite{2023_ICCV_ProGrad} & $\times$    & 96.5  & 83.8  & 82.5  & 59.7  & 80.6  \\
    
    MaPLe \cite{2023_CVPR_MaPLe} & $\times$    & 96.5 & 82.2	& 83.4	& 59.5	& 80.4 \\
    
    StyLIP \cite{2024_WACV_StyLIP} & $\times$    & \textbf{98.1 } & \textbf{86.9} & 84.6  & 62.0  & 82.9  \\

    SPG \cite{2024_arXiv_SPG}  & $\times$    & 97.0  & 82.4  & 83.6  & 60.1  & 80.8  \\
    
    VL2V-SD \cite{2024_CVPR_VL2V-ADiP}  & $\times$    & 96.7  & 83.3  & 87.4  & 62.8  & 82.5  \\
    
    CLIPood \cite{2023_ICML_CLIPood} & $\times$    & 97.3  & 85.0  & 87.0  & \textbf{63.5} & \textbf{83.2} \\
    
    Diverse Text Prompts \cite{2025_CVPR_Diverse-Text-Prompt_Domain-Generalization-in-CLIP-via-Learning-with-Diverse-Text-Prompts} & $\times$    & 97.0  & 84.8  & \textbf{87.7} & 63.1  & 83.1  \\

    DUPRG \cite{2022_arXiv_DUPRG} & $\checkmark$    & 97.1  & 83.9  & 83.6  & 59.6  & 81.1  \\
    
    PromptStyler \cite{2023_ICCV_PromptStyler} & $\checkmark$   & 97.2  & 82.9  & 83.6  & 59.4  & 80.8  \\
    
    DPStyler \cite{2024_arXiv_DPStyler} & $\checkmark$   & 97.1  & 84.0  & 82.8  & 58.9  & 80.7  \\
    
    PromptTA \cite{2024_arXiv_PromptTA} & $\checkmark$   & 97.3  & 83.6  & 82.9  & 59.4  & 80.8  \\
    \bottomrule
    
    \end{tabular}
    }
  \label{tab:DG-performance}
\end{table}

\subsection{Discussion}
\label{sec_DG_Discussion}

Domain generalization (DG) has witnessed a growing number of CLIP-based studies in recent years.  
To examine the current status and emerging trends, we refer to the results in Table~\ref{tab:DG-performance}, where the Office-Home~\cite{2017_CVPR_Office-Home-dataset_Deep-Hashing-Network-for-Unsupervised-Domain-Adaptation}, PACS~\cite{2017_ICCV_PACS-dataset_Deep-Broader-and-Artier-Domain-Generalization}, VLCS~\cite{2013_ICCV_VLCS-dataset_Unbiased-Metric-Learning--On-the-Utilization-of-Multiple-Datasets-and-Web-Images-for-Softening-Bias}, and DomainNet~\cite{2019_ICCV_DomainNet-dataset_Moment-Matching-for-Multi-source-Domain-Adaptation} benchmarks are the most widely adopted datasets for evaluating cross-domain generalization.  
Several consistent tendencies can be observed:

\textbf{\textit{(i) Prompt optimization has evolved from single, handcrafted templates to diversified and deeper mechanisms.}} Early prompt tuning (e.g., class-agnostic or class-conditional prompts) yields modest but stable gains over zero-shot CLIP; coupling prompts across text \emph{and} image pathways and injecting prompts at deeper layers further improves robustness; recent prompt diversification/stylization systematically enlarges coverage of latent domain shifts, delivering the strongest gains on harder benchmarks.

\textbf{\textit{(ii) Methods increasingly exploit CLIP’s intrinsic vision–language alignment for semantic-level generalization rather than explicit domain alignment.}} By leveraging semantic anchoring via improved training objectives, constraints, or distillation, recent approaches transfer cross-modal knowledge more effectively to the visual encoder. Hybrid strategies that combine semantic cues with lightweight alignment remain present, but the dominant trend is semantics-first.

\textbf{\textit{(iii) With a fixed ViT-B/16 backbone, the main improvements stem from better use of multimodal semantics and prompt design, not architectural complexity.}} PACS \cite{2017_ICCV_PACS-dataset_Deep-Broader-and-Artier-Domain-Generalization} and VLCS \cite{2013_ICCV_VLCS-dataset_Unbiased-Metric-Learning--On-the-Utilization-of-Multiple-Datasets-and-Web-Images-for-Softening-Bias} are near-saturated, while Office-Home \cite{2017_CVPR_Office-Home-dataset_Deep-Hashing-Network-for-Unsupervised-Domain-Adaptation} and DomainNet \cite{2019_ICCV_DomainNet-dataset_Moment-Matching-for-Multi-source-Domain-Adaptation} reveal the benefits of semantic utilization and prompt diversity; performance differences primarily reflect how well methods capitalize on CLIP’s cross-modal priors.

\textbf{\textit{(iv) Source-free (SF) variants offer practical trade-offs.}} Without access to source data, SF methods employ stylized or synthetic prompts/adapters to approximate domain diversity and reach competitive mid–high performance, yet they typically trail the best non-SF counterparts on the most challenging benchmarks.

\textbf{\textit{Overall,}} CLIP-powered DG is trending toward \textit{knowledge-driven, semantically grounded} generalization: \textit{efficient and diversified prompt learning} combined with \textit{cross-modal transfer} constitutes the primary path to improved robustness under a fixed backbone, suggesting future work on more realistic shift evaluations and narrowing the gap between \textit{SA} and \textit{SF}.

\section{Domain Adaptation (DA)}
\label{sec_DA}

CLIP-based DA aims to achieve reliable adaptation from labeled source domains (except SFF scenario) to unlabeled or sparsely labeled targets under distribution shifts by exploiting CLIP’s transferable vision–language representations.
This section reviews recent advances from two perspectives: source-available (SA) (\S\ref{sec_DA_SA}) and source-free (SF) (\S\ref{sec_DA_SF}), followed by comparative analyses and a consolidated discussion (\S\ref{sec_DA_Discussion}) of emerging trends.

\subsection{Source-Available (SA)}
\label{sec_DA_SA}

This subsection further organizes source-available (SA) adaptation methods into single-source (\S\ref{sec_DA_SA_SS}) and multi-source (\S\ref{sec_DA_SA_MS}) settings, outlining their methodological characteristics and representative frameworks across different label-space scenarios.

\subsubsection{Single-Source (SS)}
\label{sec_DA_SA_SS}

\paragraph{\textbf{Single-Source Closed-Set Unsupervised Domain Adaptation (SS-CSUDA)}}
\label{sec_DA_SA_SS_CS}

\textit{In SS-CSUDA, the label space of the target domain $ \mathcal{Y}_t $ is identical to that of the source domain $ \mathcal{Y}_s $, i.e., $ \mathcal{Y}_t = \mathcal{Y}_s $. In this scenario, a model $ f: \mathbb{R}^{H \times W} \mapsto \mathbb{R}^{C} $ is trained on a labeled source domain $ \mathcal{D}_s = \{(x_i, y_i)\}_{i=1}^{N_s} $ together with a target domain $ \mathcal{D}_t = \{x_j\}_{j=1}^{N_t} $ consisting of unlabeled samples, with the objective of adapting the model to the target domain.
}

Existing studies enhance single-source adaptation by leveraging labeled source data to transfer multimodal knowledge and mitigate distribution shifts between domains. 
These methods can be broadly categorized into four groups: 
\textit{(i) domain-aware and dual-branch prompt tuning} that integrates domain-specific or adversarial prompt mechanisms for better adaptation \cite{2023_ICCV_PADCLIP, 2023_ICCVW_AD-CLIP, 2024_AAAI_PDA, 2024_WACV_PTT-VFR, 2025_IJCV_MAwLLM_Multi-modal-Prompt-Alignment-with-Fine-grained-LLM-Knowledge-for-Unsupervised-Domain-Adaptation, 2025_KBS_PDbDa_PDbDa=Prompt-Tuned-Dual-branch-Framework-for-Unsupervised-Domain-Adaptation, 2025_TIP_ADAPT_When-Adversarial-Training-Meets-Prompt-Tuning=Adversarial-Dual-Prompt-Tuning-for-Unsupervised-Domain-Adaptation}, 
\textit{(ii) cross-modal alignment and bridging mechanisms} that align visual and textual embeddings for semantic consistency \cite{2024_CVPR_DAMP, 2024_CVPR_UniMoS, 2025_TGRS_PADA-Net_Prompt-Integrated-Adversarial-Unsupervised-Domain-Adaptation-for-Scene-Recognition, 2025_TOMM_PIMA_Prompt-Based-Invertible-Mapping-Alignment-for-Unsupervised-Domain-Adaptation, 2025_TOMM_UTISA_Unified-Text-image-Space-Alignment-with-Cross-Modal-Prompting-in-CLIP-for-UDA}, 
\textit{(iii) distribution alignment and regularization strategies} that refine pseudo-labels and enforce feature consistency across domains \cite{2024_arXiv_CLIP-Div, 2024_IJCNN_DACR, 2024_TCSVT_CMKD, 2025_WACV_SWG_Combining-Inherent-Knowledge-of-Vision-language-Models-with-Unsupervised-Domain-Adaptation-through-Strong-weak-Guidance, 2025_CVPR_CRPL_Preserving-Clusters-in-Prompt-Learning-for-Unsupervised-Domain-Adaptation}, 
and \textit{(iv) fuzzy system–enhanced adaptation} that leverages fuzzy logic to handle uncertainty in cross-domain learning \cite{2024_FUZZ-IEEE_VLM-TSK-DA, 2024_TOFS_FUZZLE}.

\textbf{\textit{(i) Domain-aware and dual-branch prompt tuning.}}
PADCLIP~\cite{2023_ICCV_PADCLIP} incorporates domain names into prompts and mitigates catastrophic forgetting through adaptive debiasing, adjusting causal inference with momentum and forgetting measurement.
Similarly,
AD-CLIP~\cite{2023_ICCVW_AD-CLIP} learns domain-invariant and class-generic prompt tokens, including domain, image, and class tokens, based on visual features.
Likewise,
DAPrompt~\cite{2023_TNNLS_DAPrompt} utilizes domain-specific prompts and incorporates a dynamic mechanism to adapt the classifier to each domain.
Subsequently,
PDA~\cite{2024_AAAI_PDA} integrates domain knowledge into prompt learning using a two-branch prompt tuning framework with base and alignment branches.
Meanwhile,
PTT-VFR~\cite{2024_WACV_PTT-VFR} introduces prompt task-dependent tuning and visual feature refinement, using domain-aware pseudo-labeling and zero-shot predictions for efficient adaptation.
With the rapid advancement of Large Language Models (LLMs),
MAwLLM~\cite{2025_IJCV_MAwLLM_Multi-modal-Prompt-Alignment-with-Fine-grained-LLM-Knowledge-for-Unsupervised-Domain-Adaptation} is the first to incorporate fine-grained LLM knowledge into prompt learning, leveraging hybrid-modal guidance for improved visual–textual alignment and stronger target adaptation.
Recently, 
PDbDa \cite{2025_KBS_PDbDa_PDbDa=Prompt-Tuned-Dual-branch-Framework-for-Unsupervised-Domain-Adaptation} introduces a prompt-tuned dual-branch framework that integrates a foundation branch with prompt knowledge constraint and an adaptation branch with domain-aware feature tuning.
In parallel, 
ADAPT~\cite{2025_TIP_ADAPT_When-Adversarial-Training-Meets-Prompt-Tuning=Adversarial-Dual-Prompt-Tuning-for-Unsupervised-Domain-Adaptation} employs adversarial dual-prompt tuning, where text and visual prompts are jointly optimized to align source and target domains via adversarial cross-modal interaction.

\textbf{\textit{(ii) Cross-modal alignment and bridging mechanisms.}}
DAMP~\cite{2024_CVPR_DAMP} aligns visual and textual embeddings to leverage domain-invariant semantics, using image context to prompt the language branch in a domain-agnostic, instance-conditioned manner.
Subsequently,
UniMoS~\cite{2024_CVPR_UniMoS} separates CLIP-extracted visual features into vision and language components, which are trained independently and then aligned using a modality discriminator.
In addition,
PADA-Net~\cite{2025_TGRS_PADA-Net_Prompt-Integrated-Adversarial-Unsupervised-Domain-Adaptation-for-Scene-Recognition} integrates prompt learning with an interactive bridging mechanism for cross-modal exchange, dual Meta-nets to improve feature discriminability, and optimal transport with game-theoretic strategies for visual–linguistic alignment.
Meanwhile,
PIMA~\cite{2025_TOMM_PIMA_Prompt-Based-Invertible-Mapping-Alignment-for-Unsupervised-Domain-Adaptation} leverages invertible mapping to preserve domain topology during prompt-based alignment and introduces cross-modal implicit contrastive learning to regularize non-informative features for semantic consistency.
Recently,
UTISA~\cite{2025_TOMM_UTISA_Unified-Text-image-Space-Alignment-with-Cross-Modal-Prompting-in-CLIP-for-UDA} constructs domain- and layer-specific image prompts via cross-modal prompting to align textual and visual embeddings across domains, and uses domain-aware multi-layer fusion to enhance image representations and their alignment with domain-specific text prompts.

\textbf{\textit{(iii) Distribution alignment and regularization strategies.}}
CLIP-Div~\cite{2024_arXiv_CLIP-Div} aligns source and target domains by CLIP's domain-agnostic distribution, with two new divergence losses: absolute and relative loss to enhance alignment. Different from existing pseudo-labeling-related methods
~\cite{2017_ICML_Deep-Transfer-Learning-with-Joint-Adaptation-Networks, 2018_NeurIPS_Conditional-Adversarial-Domain-Adaptation, 2018_CVPR_Collaborative-and-Adversarial-Network-for-Unsupervised-Domain-Adaptation, 2018_ECCV_Unsupervised-Domain-Adapation-for-Semantic-Segmentation-via-Class-balanced-Self-training, 2019_arXiv_Pseudo-labeling-Curriculum-for-Unsupervised-Domain-Adaptation, 2020_NeurIPS_FixMatch, 2020_ECCV_Instance-Adaptative-Self-training-for-Unsupervised-Domain-Adaptation, 2021_NeurIPS_Cycle-Self-training-for-Domain-Adaptation, 2021_NeurIPS_FlexMatch, 2022_CVPR_Cross-domain-Adaptative-Teacher-for-Object-Detection, 2022_CVPR_Semi-supervised-Learning-of-Semantic-Correspondence-with-Pseudo-labels}, it employs a language-guided pseudo-labeling strategy designed to calibrate target pseudo-labels, which enhances the model’s generalization on the target domain through self-training.
Likewise,
DACR~\cite{2024_IJCNN_DACR} optimizes prompts and image adapters with consistency regularization, improving generalization and domain-specific feature learning through pseudo-label consistency across augmented views.
In addition,
CMKD~\cite{2024_TCSVT_CMKD} incorporates residual sparse training to greatly reduce parameter storage requirements and enhance deployment efficiency.
Meanwhile,
SWG~\cite{2025_WACV_SWG_Combining-Inherent-Knowledge-of-Vision-language-Models-with-Unsupervised-Domain-Adaptation-through-Strong-weak-Guidance} integrates CLIP’s zero-shot knowledge into domain alignment through a strong–weak guidance scheme, where strong guidance selects high-confidence target samples and weak guidance distills softened predictions to preserve VLM semantics during adaptation.
Furthermore,
CRPL~\cite{2025_CVPR_CRPL_Preserving-Clusters-in-Prompt-Learning-for-Unsupervised-Domain-Adaptation} also performs experiments under this setting to verify its generalization capability.

\textbf{\textit{(iv) Fuzzy system–enhanced adaptation.}}
VLMTSK-DA~\cite{2024_FUZZ-IEEE_VLM-TSK-DA} enhances VLMs through the integration of Takagi-Sugeno-Kang (TSK) fuzzy systems. Inspired by some fuzzy systems methods
~\cite{2019_TOFS_Fuzzy-Multiple-Source-Transfer-Learning, 2019_TOFS_Transfer-Representation-Learning-with-TSK-Fuzzy-System, 2020_TOFS_Multisource-Heterogeneous-Unsupervised-Domain-Adaptation-via-Fuzzy-Relation-Neural-Networks, 2023_TOFS_Source-Free-Multidomain-Adaptation-with-Fuzzy-Rule-Based-Deep-Nueral-Networks, 2023_FUZZ-IEEE_Attention-Bridging-TS-Fuzzy-Rules-for-Universal-Multi-Domain-Adaptation-without-Source-Data},
It uses the TSK system as an image adapter to manage uncertainty, combining image features residually for optimized performance. Through prompt learning, it synchronizes visual and textual updates for global optimization and applies fuzzy C-means loss to align target data with source clusters, reducing distribution gaps.
Similarly,
FUZZLE~\cite{2024_TOFS_FUZZLE} integrates fuzzy techniques
~\cite{2018_TOFS_Fuzzy-Rule-based-Domain-Adaptation-in-Homogeneous-and-Heterogeneous-Spaces, 2018_TOFS_Unsupervised-Heterogeneous-Domain-Adaptation-via-Shared-Fuzzy-Equivalence-Relations, 2020_TOFS_Multisource-Heterogeneous-Unsupervised-Domain-Adaptation-via-Fuzzy-Relation-Neural-Networks, 2020_TOFS_Low-Rank-Tensor-Regularized-Fuzzy-Clustering-for-Multiview-Data, 2021_TOFS_Fuzzy-Multioutput-Transfer-Learning-for-Regression, 2023_TOFS_Source-free-Multidomain-Adaptation-with-Fuzzy-Rule-Based-Deep-Neural-Netowrks}
into prompt learning for the first time within the vision–language model domain. It employs domain-specific prompt learning, fuzzy C-means, and instance-level fuzzy vectors to align prompts with cluster centers. A KL divergence loss with a fuzzification factor reduces cross-domain gaps during training.

\paragraph{\textbf{Single-Source Open-Set Unsupervised Domain Adaptation (SS-OSUDA)}}
\label{sec_DA_SA_SS_OS}

\textit{In SS-OSUDA, the label space of the target domain $ \mathcal{Y}_t $ completely includes the classes from the source domain $ \mathcal{Y}_s $ and introduces additional novel classes, i.e., $ \mathcal{Y}_s \cap \mathcal{Y}_t \neq \emptyset $ and $ \mathcal{Y}_s \subset \mathcal{Y}_t $. In this scenario, a model $ f: \mathbb{R}^{H \times W} \mapsto \mathbb{R}^{C} $ is trained on a labeled source domain $ \mathcal{D}_s = \{(x_i, y_i)\}_{i=1}^{N_s} $ together with a target domain $ \mathcal{D}_t = \{x_j\}_{j=1}^{N_t} $ consisting of unlabeled samples, with the objective of adapting the model to the target domain.
}

Existing studies extend CLIP’s capability to recognize both known and novel classes in the target domain while mitigating performance drops under domain shifts. They mainly achieve this via entropy-based calibration, prompt disentanglement, and domain-aware bias learning to enhance open-set robustness with unlabeled targets.

\citet{2023_arXiv_2025_CVIU_ODA-with-CLIP} proposes ODA with CLIP, and it calculates the entropy of the outputs of the ODA model and the predictions of CLIP on the target domain to identify known and unknown samples.
Subsequently,
PromptDIV~\cite{2024_ICIP_PromptDIV} decouples domain invariance and variance through tailored prompt learning, using one-vs-all text-based clustering for domain-unbiased pseudo-labeling, domain-specific prompts to disentangle invariant and variant representations, and semi-supervised affinity contrastive learning to reinforce intra-class consistency.
Meanwhile,
COSMo~\cite{2024_arXiv_COSMo} employs source-guided prompt learning with a domain-specific bias network and separate prompts for known and unknown classes to achieve domain-agnostic generalization across multiple targets.

\subsubsection{Multi-Source (MS)}
\label{sec_DA_SA_MS}

\paragraph{\textbf{Multi-Source Closed-Set Unsupervised Domain Adaptation (MS-CSUDA)}}
\label{sec_DA_SA_MS_CS}

\textit{In MS-CSUDA, the label space of the target domains $\mathcal{Y}_t$ is identical to that of the source domains $\mathcal{Y}_s$, i.e., $\mathcal{Y}_t = \mathcal{Y}_s$. A model $f: \mathbb{R}^{H \times W} \mapsto \mathbb{R}^{C}$ is trained on multiple labeled source domains $ \mathcal{D}_s = \{D_1, D_2, \ldots, D_S\} $ together with a target domain $ \mathcal{D}_t = \{x_j\}_{j=1}^{N_t} $ consisting of unlabeled samples, with the objective of adapting the model to the target domain.}

Existing studies enhance multi-source adaptation by aggregating complementary knowledge from diverse sources to improve generalization on unlabeled targets.
They primarily employ prompt learning and cross-modal alignment to unify domain semantics, reduce distribution gaps, and achieve robust adaptation under closed-set conditions.

MPA~\cite{2023_NeurIPS_MPA} designs a prompt learning-based method which trains individual prompts for each domain pair and aligns them using an auto-encoder, with an latent subspace tuning strategy for efficient adaptation to target domains.
Meanwhile,
LanDA~\cite{2024_arXiv_LanDA} introduces a language-guided method that transfers knowledge from multiple source domains to a target domain using only textual descriptions, preserving task-relevant information without requiring target domain images.
In parallel,
MSDPL~\cite{2024_OpenReview_MSDPL} is also a prompt learning methodology that enhances VLMs using domain prompts. It proposes "domain-aware mixup," improving domain prompt learning. It focuses on both domain and class prompts to enhance performance in few-shot adaptation tasks.
Recently,
VAMP~\cite{2025_AAAI_VAMP_Vision-aware-Multimodal-Prompt-Tuning-for-Uploadable-Multi-source-Few-shot-Domain-Adaptation} uses vision-aware prompts to produce domain-guided text prompts that enhance semantic discriminability, trained in a decentralized framework with cross-modal alignment and lightweight regularizations.
CRPL~\cite{2025_CVPR_CRPL_Preserving-Clusters-in-Prompt-Learning-for-Unsupervised-Domain-Adaptation} reinforces pseudo-label learning by combining source-enhanced predictions with clustering-preserving regularization based on Wasserstein distance, effectively aligning visual and textual embeddings for more discriminative target prompts.

\begin{figure}[t]
    \centering
    \includegraphics[width=0.75\linewidth]{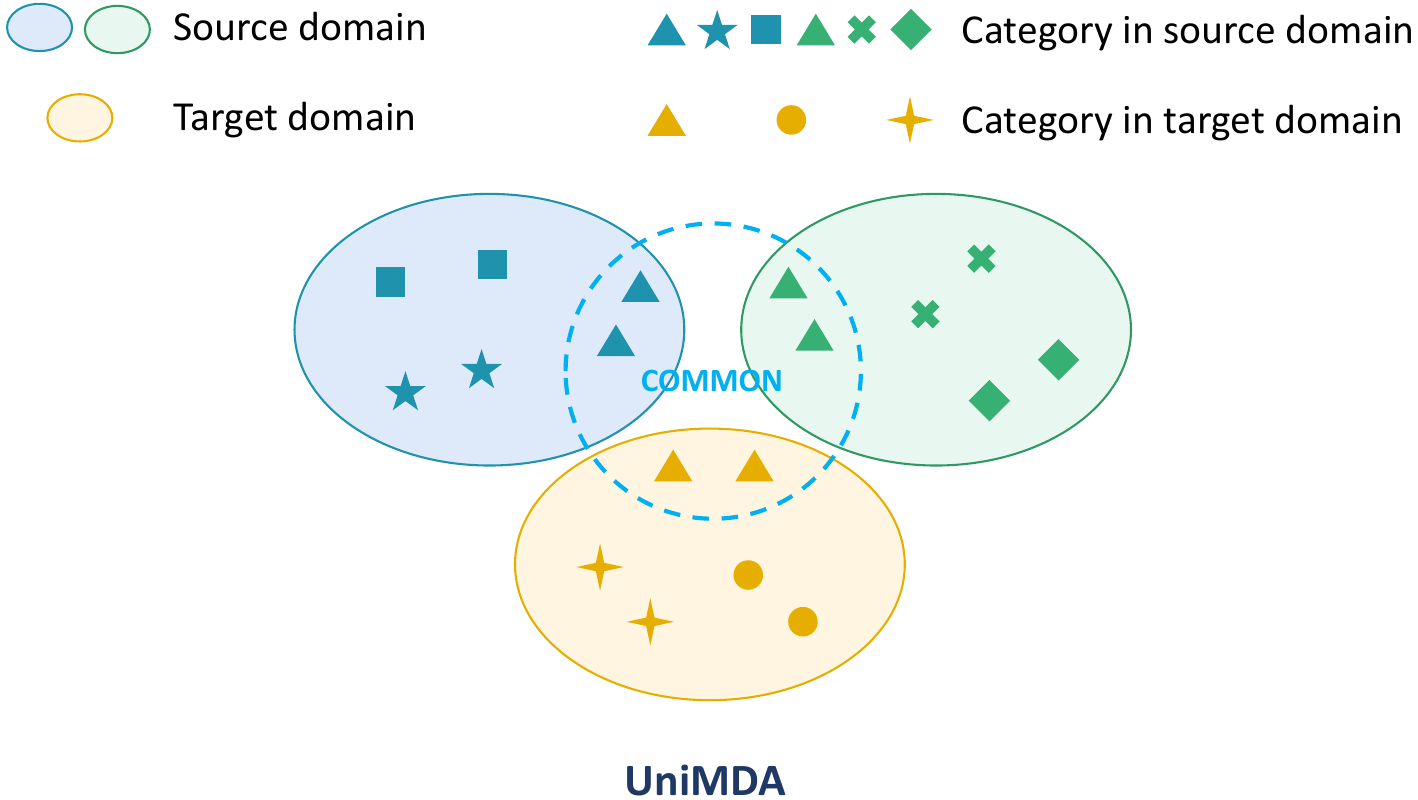}
    \caption{The scenario of Multi-Source Open-Partial-Set Unsupervised Domain Adaptation (MS-OPSUDA) a.k.a. Universal Multi-Source Domain Adaptation (UniMDA).}
    \label{fig:UniMDA}
\end{figure}

\paragraph{\textbf{Multi-Source Open-Partial-Set Unsupervised Domain Adaptation (MS-OPSUDA) a.k.a. Universal Multi-Source Domain Adaptation (UniMDA)}}
\label{sec_DA_SA_MS_OPS}

\textit{In MS-OPSUDA (UniMDA), the label space of the target domain $ \mathcal{Y}_t $ partially overlaps with the label space of the source domains $ \mathcal{Y}_s = \{\mathcal{Y}_1, \mathcal{Y}_2, \ldots, \mathcal{Y}_S\} $, i.e., $ \mathcal{Y}_s \cap \mathcal{Y}_t \neq \emptyset $, $ \mathcal{Y}_t \nsubseteq \mathcal{Y}_s $, and $ \mathcal{Y}_s \nsubseteq \mathcal{Y}_t $ (illustrated in Fig.~\ref{fig:UniMDA}). In this scenario, a model $ f: \mathbb{R}^{H \times W} \mapsto \mathbb{R}^{C} $ is trained on multiple labeled source domains $ \mathcal{D}_s = \{D_1, D_2, \ldots, D_S\} $ together with a target domain $ \mathcal{D}_t = \{x_j\}_{j=1}^{N_t} $ consisting of unlabeled samples, with the objective of adapting the model to the target domain.}

Existing studies in this scenario aim to address both domain and class shifts by learning discriminative representations for shared classes while identifying unknown samples in the target domain.
However, CLIP-specific work remains limited, and existing non-CLIP methods such as UMAN~\cite{2022_Pattern-Recognition_Universal-Multi-source-Domain-Adaptation-for-Image-Classification} and HyMOS~\cite{2022_WACV_Distance-based-Hyperspherical-Classification-for-Multi-source-Open-set-Domain-Adaptation} primarily investigate unified adaptation across heterogeneous sources rather than leveraging vision–language priors.

As a CLIP-based attempt,
SAP-CLIP~\cite{2024_Signal-Processing-Letters_SAP-CLIP} proposes a semantic-aware adaptive prompt method that handles domain and class shifts using learnable prompts and employs a dynamic margin loss to better separate unknown from known samples.

\subsection{Source-Free (SF)}
\label{sec_DA_SF}

This section is structured based on two main categories: Source-Fully-Free (SFF) (\S\ref{sec_DA_SF_SFF}: \S\ref{sec_DA_SF_SFF_CS}, \S\ref{sec_DA_SF_SFF_PS}, \S\ref{sec_DA_SF_SFF_OS}, \S\ref{sec_DA_SF_SFF_OPS}) and Source-Data-Free (SDF) (\S\ref{sec_DA_SF_SDF}: \S\ref{sec_DA_SF_SDF_CS}, \S\ref{sec_DA_SF_SDF_PS}, \S\ref{sec_DA_SF_SDF_OS}). Each category addresses distinct strategies and challenges in training models under these conditions.

\subsubsection{Source-Fully-Free (SFF)}
\label{sec_DA_SF_SFF}

\paragraph{\textbf{Source-Fully-Free Closed-Set Unsupervised Domain Adaptation (SFF-CSUDA) a.k.a. Closed-Set Unsupervised Fine-Tuning (CS-UFT)}}
\label{sec_DA_SF_SFF_CS}

\textit{Source-Fully-Free Closed-Set Unsupervised Domain Adaptation (SFF-CSUDA), also known as Closed-Set Unsupervised Fine-Tuning (CS-UFT), refers to a scenario where the label space of the target domain $ \mathcal{Y}_t $ is identical to that of the predetermined list of known classes $ \mathcal{Y}_s $, i.e., $ \mathcal{Y}_t = \mathcal{Y}_s $. In this scenario, the model $ f: \mathbb{R}^{H \times W} \mapsto \mathbb{R}^{C} $ is adapted to a target domain $ \mathcal{D}_t = \{x_j\}_{j=1}^{N_t} $ consisting of unlabeled samples, 
\textcolor{revision-color_1}{
with no access to either data from the source domain $ \mathcal{D}_s $ or any source domain knowledge (e.g., source-trained models or statistics on the source domain).}
}

Existing studies primarily rely on pseudo-label refinement, cross-modal consistency, and parameter-efficient tuning to exploit the knowledge already embedded in CLIP. These methods strengthen the alignment between textual prototypes and target features while progressively improving label quality.

UPL~\cite{2022_arXiv_UPL} is the first work to introduce unsupervised learning into prompt learning of VLM. The method introduces several techniques, such as a top-K pseudo-labeling strategy, pseudo label ensemble, and prompt representation ensemble.
Similarly,
POUF~\cite{2023_ICML_POUF} aligns prototypes and target data in the latent space using transport-based distribution alignment and mutual information maximization. 
In parallel,
ReCLIP~\cite{2024_WACV_ReCLIP} learns a projection space to correct misaligned visual–text embeddings and produce pseudo labels, then applies cross-modal self-training to iteratively refine the encoders and further reduce domain gap and misalignment.
As another approach,
LaFTer~\cite{2024_NeurIPS_LaFTer} is a label-free, parameter-efficient cross-modal transfer method using only 0.4\% of parameters. It fine-tunes a VLM to improve target-class recognition with automatically generated texts and an unlabeled image set.
Also,
CPL~\cite{2024_ICML_CPL} introduces candidate pseudolabel learning for fine-tuning VLMs with unlabeled data. It refines pseudo labels by constructing a confidence score matrix and leveraging both intra- and inter-instance selection to improve true label identification.
Moreover,
DPA~\cite{2024_arXiv_DPA} generates accurate pseudo-labels by combining outputs from two prototypes and ranking them to reduce noise, especially early in training. It also aligns textual prototypes with image prototypes to address visual-textual misalignment.
Likewise,
~\citet{2024_ICML_UEO} proposes Universal Entropy Optimization (UEO) and it leverages sample-level confidence to optimize entropy and jointly tunes textual prompts and visual affine transformations in CLIP.
Additionally,
TFUP~\cite{2024_arXiv_TFUP} enhances features with similarity-based predictions and selects samples using confidence and prototype scores.
Furthermore,
\citet{2024_ICIP_Rethinking-Domain-Adaptation-and-Generalization-in-the-Era-of-CLIP} shows that domain priors enhance CLIP's zero-shot performance and introduces a benchmark with a knowledge disentangling strategy for label-free multi-source generalization.


\begin{table*}[htbp]
\centering
  
\caption{\textcolor{revision-color_1}{Performance (accuracy, (\%)) comparison of representative DA methods on OfficeHome~\cite{2017_CVPR_Office-Home-dataset_Deep-Hashing-Network-for-Unsupervised-Domain-Adaptation} . All ResNet-based methods use ResNet-50 as the backbone, and ViT-B methods use ViT-B/16 by default, the -B32 suffix denote ViT-B/32. (We report the result from their related original paper.)}}
\renewcommand{\arraystretch}{1.3}
\scalebox{0.6}{
    \begin{tabular}{>{\centering\arraybackslash}p{3.2cm} | >{\centering\arraybackslash}p{1cm} | >{\centering\arraybackslash}p{1.1cm} | >{\centering\arraybackslash}p{1.1cm} | >{\centering\arraybackslash}p{1cm} >{\centering\arraybackslash}p{1cm} >{\centering\arraybackslash}p{1cm} >{\centering\arraybackslash}p{1cm} >{\centering\arraybackslash}p{1cm} >{\centering\arraybackslash}p{1cm} >{\centering\arraybackslash}p{1cm} >{\centering\arraybackslash}p{1cm} >{\centering\arraybackslash}p{1cm} >{\centering\arraybackslash}p{1cm} >{\centering\arraybackslash}p{1cm} >{\centering\arraybackslash}p{1cm} | >{\centering\arraybackslash}p{1.7cm}}
    \toprule
    method & SDF   & SFF   & E     & C-A   & P-A   & R-A   & A-C   & P-C   & R-C   & A-P   & C-P   & R-P   & A-R   & C-R   & P-R   & AVG \\
    \midrule
    
    CLIP \cite{2021_ICML_CLIP}  & N/A   & N/A   & \multirow{16}[2]{*}{ResNet} & 71.7  & 71.7  & 71.7  & 51.7  & 51.7  & 51.7  & 81.5  & 81.5  & 81.5  & 82.3  & 82.3  & 82.3  & 71.8  \\
    
    DAPL (DAPrompt) \cite{2023_TNNLS_DAPrompt} & $\times$     & $\times$     &       & 74.4  & 74.5  & 75.2  & 54.1  & 54.6  & 54.7  & 84.3  & 83.7  & 83.8  & 84.8  & 85.0  & 84.8  & 74.5  \\

    CLIP-DIV \cite{2024_arXiv_CLIP-Div} & $\times$     & $\times$     &       & 73.9  & 72.8  & 73.3  & 57.2  & 58.6  & 59.9  & 80.4  & 80.7  & 81.9  & 82.9  & 81.1  & 83.5  & 73.9  \\

    AD-CLIP \cite{2023_ICCVW_AD-CLIP}   & $\times$     & $\times$     &       & 76.1  & 76.7  & 76.8  & 55.4  & 56.1  & 56.1  & 85.2  & 85.8  & 85.5  & 85.6  & 86.2  & 85.4  & 75.9  \\
    
    PADCLIP \cite{2023_ICCV_PADCLIP} & $\times$     & $\times$     &       & 77.8  & 76.3  & 78.1  & 57.5  & 59.2  & 60.2  & 84.0  & 85.5  & 86.7  & 83.8  & 84.7  & 85.4  & 76.6  \\
    
    UniMos \cite{2024_CVPR_UniMoS} & $\times$     & $\times$     &       & 75.2  & 75.4  & 76.9  & 59.5  & 58.4  & 59.5  & 89.4  & 89.6  & 89.7  & 86.9  & 86.8  & 87.2  & 77.9  \\
    
    DAMP \citep{2024_CVPR_DAMP}  & $\times$     & $\times$     &       & 76.6  & 76.3  & 77.0  & 59.7  & 59.6  & 61.0  & 88.5  & 88.9  & 89.9  & 86.8  & 87.0  & 87.1  & 78.2  \\
    
    CMKD \cite{2024_TCSVT_CMKD} & $\times$     & $\times$     &       & 74.4  & 75.9  & 79.1  & 65.9  & 64.4  & \textbf{67.2 } & 86.6  & 87.7  & 90.0  & 87.3  & 85.8  & 87.9  & 79.3  \\
    
    PTT-VFR \cite{2024_WACV_PTT-VFR} & $\times$     & $\times$     &       & 77.4  & 76.5  & 75.9  & 58.1  & 58.8  & 60.4  & 85.0  & 85.0  & 86.4  & 84.5  & 84.7  & 85.7  & 76.5  \\

    DACR \cite{2024_IJCNN_DACR} & $\times$     & $\times$     &       & 75.8  & 74.5  & 75.2  & 54.6  & 54.1  & 54.9  & 84.6  & 84.2  & 84.0  & 85.1  & 85.1  & 85.2  & 74.8  \\
    
    FUZZLE \cite{2024_TOFS_FUZZLE} & $\times$     & $\times$     &       & 75.0  & 74.0  & 75.1  & 55.9  & 56.2  & 56.2  & 84.6  & 84.8  & 86.4  & 85.5  & 84.3  & 85.1  & 75.3  \\
    
    VLM-TSK-DA \cite{2024_FUZZ-IEEE_VLM-TSK-DA} & $\times$     & $\times$     &       & 74.9  & 73.4  & 74.2  & 55.0  & 54.1  & 54.2  & 86.3  & 86.3  & 86.2  & 86.0  & 84.7  & 83.8  & 74.9  \\
    
    DIFO \cite{2024_CVPR_DIFO} & $\checkmark$ & $\times$     &       & 79.5  & 78.3  & 80.0  & 62.6  & 63.4  & 63.3  & 87.5  & 87.9  & 87.7  & 87.1  & 87.4  & 88.1  & 79.4  \\
    
    CDBN \cite{2024_arXiv_CDBN} & $\checkmark$ & $\times$     &       & 77.0  & 76.6  & 76.9  & 59.2  & 59.1  & 59.9  & 89.4  & 89.0  & 88.9  & 87.1  & 87.0  & 87.3  & 78.1  \\
    
    BBC \cite{2024_Signal-Image-and-Video-Processing_BBC}  & $\checkmark$ & $\times$     &       & 76.4  & 76.7  & 78.1  & \textbf{67.5 } & \textbf{66.7 } & 66.2  & 87.4  & 89.0  & 89.3  & 87.1  & 87.3  & 87.3  & 79.9  \\
    
    ProDe \cite{2025_ICLR_ProDe_Proxy-Denoising-for-Source-Free-Domain-Adaptation} & $\checkmark$ & $\times$     &       & \textbf{81.1 } & \textbf{79.8 } & \textbf{80.9 } & 64.0  & 65.4  & 65.5  & \textbf{90.0 } & \textbf{90.1 } & \textbf{90.2 } & \textbf{88.3 } & \textbf{88.6 } & \textbf{89.0 } & \textbf{81.1 } \\
    
    \midrule
    
    CLIP \cite{2021_ICML_CLIP} & N/A   & N/A   & \multirow{17}[2]{*}{ViT-B} & 82.9  & 82.9  & 82.9  & 67.8  & 67.8  & 67.8  & 89.0  & 89.0  & 89.0  & 89.8  & 89.8  & 89.8  & 82.4  \\
    
    CLIP-DIV \cite{2024_arXiv_CLIP-Div} & $\times$     & $\times$     &       & 83.0  & 81.5  & 81.8  & 71.4  & 73.0  & 73.9  & 89.5  & 89.9  & 90.2  & 89.4  & 89.0  & 89.4  & 83.5  \\
    
    DAPL (DAPrompt) \cite{2023_TNNLS_DAPrompt}  & $\times$     & $\times$     &       & 85.2  & 85.1  & 85.3  & 70.7  & 70.7  & 70.4  & 91.0  & 91.0  & 91.4  & 90.9  & 91.0  & 90.9  & 84.4  \\
    
    PADCLIP \citep{2023_ICCV_PADCLIP} & $\times$     & $\times$     &       & 86.7  & 86.0  & 86.9  & 76.4  & 74.5  & 79.1  & 90.6  & 92.3  & 93.1  & 90.8  & 92.0  & 91.5  & 86.7  \\
    
    AD-CLIP \cite{2023_ICCVW_AD-CLIP} & $\times$     & $\times$     &       & 85.4  & 86.7  & 86.9  & 70.9  & 74.3  & 72.6  & 92.5  & 92.4  & 93.8  & 92.1  & 92.5  & \textbf{93.0 } & 86.1  \\
    
    PDA \cite{2024_AAAI_PDA}  & $\times$     & $\times$     &       & 86.0  & 86.0  & 86.4  & 73.5  & 73.5  & 73.0  & 91.4  & 91.6  & 92.4  & 91.3  & 91.5  & 91.7  & 85.7  \\
    
    DAMP \cite{2024_CVPR_DAMP} & $\times$     & $\times$     &       & 86.3  & 86.2  & 86.1  & 75.7  & 76.3  & 75.6  & \textbf{94.2}  & 94.2  & 94.0  & 92.0  & 91.9  & 92.4  & 87.1  \\
    
    MAwLLM \cite{2025_IJCV_MAwLLM_Multi-modal-Prompt-Alignment-with-Fine-grained-LLM-Knowledge-for-Unsupervised-Domain-Adaptation} & $\times$     & $\times$     &       & 87.3  & \textbf{87.8 } & 87.6  & 76.9  & 77.2  & 77.5  & 94.0  & 93.9  & 94.0  & \textbf{92.7}  & \textbf{92.7} & 92.8  & 87.9  \\
    
    CMKD \cite{2024_TCSVT_CMKD} & $\times$     & $\times$     &       & 86.3  & 86.7  & \textbf{88.2 } & \textbf{79.4 } & 79.5  & \textbf{81.2 } & \textbf{94.2 } & 93.4  & \textbf{94.5 } & \textbf{92.7 } & 92.2  & 92.1  & \textbf{88.4 } \\
    
    PTT-VFR \cite{2024_WACV_PTT-VFR} & $\times$     & $\times$     &       & \textbf{87.5 } & 86.7  & 86.4  & 78.2  & \textbf{79.7 } & 79.4  & 90.4  & 91.9  & 93.5  & 91.0  & 92.3  & 90.9  & 87.3  \\
    
    CDBN \cite{2024_arXiv_CDBN} & $\checkmark$ & $\times$     &       & 85.7  & 84.9  & 85.5  & 73.9  & 73.7  & 72.7  & 94.1  & \textbf{94.3 } & 94.0  & 91.6  & 91.7  & 91.9  & 86.2  \\
    
    DIFO-B32 \cite{2024_CVPR_DIFO} & $\checkmark$ & $\times$     &       & 82.5  & 80.9  & 83.4  & 70.6  & 70.1  & 70.5  & 90.6  & 90.6  & 91.2  & 88.8  & 88.8  & 88.9  & 83.1  \\
    
    ProDe-B32 \cite{2025_ICLR_ProDe_Proxy-Denoising-for-Source-Free-Domain-Adaptation} & $\checkmark$ & $\times$     &       & 82.5  & 82.5  & 83.0  & 72.7  & 72.5  & 72.6  & 92.3  & 91.5  & 92.2  & 90.5  & 90.7  & 90.8  & 84.5  \\
    \cline{1-3} \cline{5-17}
    
    UPL \cite{2022_arXiv_UPL} & $\times$     & $\checkmark$ &       & \multicolumn{3}{c}{83.3 } & \multicolumn{3}{c}{67.7 } & \multicolumn{3}{c}{93.9 } & \multicolumn{3}{c}{90.7 } & 83.9  \\
    
    POUF \cite{2023_ICML_POUF} & $\times$     & $\checkmark$ &       & \multicolumn{3}{c}{86.2 } & \multicolumn{3}{c}{73.8 } & \multicolumn{3}{c}{92.7 } & \multicolumn{3}{c}{91.7 } & 86.1  \\
    
    RECLIP \cite{2024_WACV_ReCLIP} & $\times$     & $\checkmark$ &       & \multicolumn{3}{c}{86.1 } & \multicolumn{3}{c}{76.0 } & \multicolumn{3}{c}{93.9 } & \multicolumn{3}{c}{92.0 } & 87.0  \\
    
    TFUP-T \cite{2024_arXiv_TFUP} & $\times$     & $\checkmark$ &       & \multicolumn{3}{c}{86.0 } & \multicolumn{3}{c}{74.2 } & \multicolumn{3}{c}{93.1 } & \multicolumn{3}{c}{91.7 } & 86.3  \\
    
    \bottomrule
    \end{tabular}%
    }
  \label{tab:DA-performance}%
\end{table*}%

\paragraph{\textbf{Source-Fully-Free Partial-Set Unsupervised Domain Adaptation (SFF-PSUDA) a.k.a. Partial-Set Unsupervised Fine-Tuning (PS-UFT)}}
\label{sec_DA_SF_SFF_PS}

\textit{In SFF-PSUDA(PS-UFT), the label space of the target domain $ \mathcal{Y}_t $ is a subset of the predetermined list of known classes $ \mathcal{Y}_s $, i.e., $ \mathcal{Y}_t \subset \mathcal{Y}_s $. In this scenario, the model $ f: \mathbb{R}^{H \times W} \mapsto \mathbb{R}^{C} $ is adapted to a target domain $ \mathcal{D}_t = \{x_j\}_{j=1}^{N_t} $ consisting of unlabeled samples, 
with no access to either data from the source domain $ \mathcal{D}_s $ or any source domain knowledge (e.g., source-trained models or statistics on the source domain).
}

In this scenario, methods adapt CLIP to target domains containing only a subset of known classes by refining feature–prototype alignment and calibrating confidence for present categories. Existing studies primarily leverage the model’s inherent knowledge to maintain stable partial-set adaptation without requiring any source data.

\citet{2024_ICML_UEO} shows that Universal Entropy Optimization (UEO) excels in this scenario by effectively adapting to target domains with a subset of source classes. This method enhances classification accuracy and demonstrates robust performance despite the absence of certain classes, highlighting its potential in real-world applications.

\paragraph{\textbf{Source-Fully-Free Open-Set Unsupervised Domain Adaptation (SFF-OSUDA) a.k.a. Open-Set Unsupervised Fine-Tuning (OS-UFT)}}
\label{sec_DA_SF_SFF_OS}

\textit{In SFF-OSUDA(OS-UFT), the label space of the target domain $ \mathcal{Y}_t $ completely includes the predetermined list of known classes $ \mathcal{Y}_s $ and introduces additional novel classes, i.e., $ \mathcal{Y}_s \subset \mathcal{Y}_t $. In this scenario, the model $ f: \mathbb{R}^{H \times W} \mapsto \mathbb{R}^{C} $ is adapted to a target domain $ \mathcal{D}_t = \{x_j\}_{j=1}^{N_t} $ consisting of unlabeled samples, 
with no access to either data from the source domain $ \mathcal{D}_s $ or any source domain knowledge (e.g., source-trained models or statistics on the source domain).
}

Existing studies in this scenario adapt CLIP to target domains containing both known and novel classes by enhancing zero-shot discrimination and stabilizing confidence estimation under fully source-free conditions, enabling more reliable separation of unknown samples and recognition of known ones.

UOTA~\cite{2023_ICML-Workshop_UOTA} improves pre-trained zero-shot models using open-set unlabeled data. It enables implicit OOD detection and enhances classification for known classes with high computational efficiency by updating only a lightweight adapter.
In addition,
\citet{2024_ICML_UEO} shows that UEO excels in this scenario, enhancing classification of both known and new instances, and proving robust in dynamic environments with changing class distributions.
Recently,
CLIPXpert~\cite{2025_arXiv_CLIPXpert_Revisiting-CLIP-for-SF-OSDA=Unleashing-Zero-shot-Potential-with-Adaptive-Threshold-and-Training-free-Feature-Filtering} leverages adaptive thresholding and SVD-based feature filtering to dynamically separate known and unknown classes, fully exploiting CLIP’s zero-shot capability without additional training or source information.

\paragraph{\textbf{Source-Fully-Free Open-Partial-Set Unsupervised Domain Adaptation (SFF-OPSUDA) a.k.a. Open-Partial-Set Unsupervised Fine-Tuning (OPS-UFT)}}
\label{sec_DA_SF_SFF_OPS}

\textit{In SFF-OPSUDA(OPS-UFT), the label space of the target domain $ \mathcal{Y}_t $ partially overlaps with the predetermined list of known classes $ \mathcal{Y}_s $, and they have their own private classes, i.e., $ \mathcal{Y}_s \cap \mathcal{Y}_t \neq \emptyset $, $ \mathcal{Y}_t \nsubseteq \mathcal{Y}_s $, and $ \mathcal{Y}_s \nsubseteq \mathcal{Y}_t $. In this scenario, the model $ f: \mathbb{R}^{H \times W} \mapsto \mathbb{R}^{C} $ is adapted to a target domain $ \mathcal{D}_t = \{x_j\}_{j=1}^{N_t} $ consisting of unlabeled samples, 
with no access to either data from the source domain $ \mathcal{D}_s $ or any source domain knowledge (e.g., source-trained models or statistics on the source domain).
}

In this scenario, existing studies exploit CLIP’s zero-shot priors and entropy-based calibration to distinguish shared from private target classes, refining confidence estimation and decision boundaries to enable reliable adaptation under partial class overlap.

\citet{2024_ICML_UEO} demonstrates that UEO enhances SFF-OPSUDA by leveraging common classes between source and target domains. It improves classification of known classes and adapts well to new, unseen classes, making it effective in partial class overlap scenarios.

\begin{table*}[ht]
\centering
\caption{\textcolor{revision-color_1}{Different scenarios for DG and DA (based on methods reviewed in this survey). 
(SA: Source-Available, SF: Source-Free, SS: Single-Source, MS: Multi-Source, CS: Closed-Set, PS: Partial-Set, OS: Open-Set, OPS: Open-Partial-Set, U: Unsupervised, FS: Few-Shot, FT: Fine-Tuning.)}}
\label{tab:scenario-DG-DA}
\renewcommand{\arraystretch}{1.6}
\scalebox{0.55}{
    \begin{tabular}{cccccccccccc}
    \toprule
    
    \multicolumn{3}{c}{\multirow{3}{*}{\textbf{Scenario}}} 
        & \multicolumn{2}{c}{\textbf{\makecell{Source Domain}}} 
        & \multicolumn{3}{c}{\textbf{\makecell{Target Domain}}} 
        & \multicolumn{4}{c}{\textbf{\makecell{Category between \\Source Domain and Target Domain}}} 
        \\
        \cmidrule(lr){4-5} \cmidrule(lr){6-8} \cmidrule(lr){9-12}
    
    \multicolumn{3}{c}{} 
    & \textbf{\makecell{Single-Source\\(SS)}}
    & \textbf{\makecell{Multi-Source\\(MS)}}
    & \textbf{\makecell{Source \\Domain Data}}
    & \textbf{\makecell{Source Domain \\Knowledge}}
    & \textbf{\makecell{Target Data \\(Unlabeled)}}
    & \textbf{\makecell{Closed-Set\\(CS)}}
    & \textbf{\makecell{Partial-Set\\(PS)}}
    & \textbf{\makecell{Open-Set\\(OS)}}
    & \textbf{\makecell{Open-Partial-Set\\(OPS)}}
    \\ 
    \midrule
    
    \multirow{4}{*}{\makecell{Domain \\Generalization\\(DG)}}
    & \multirow{3}{*}{\makecell{ Source-Available (SA)}}
    & \cellcolor{mytabcolor_1}SS-CSDG (\S\ref{sec_DG_CLIP-as-backbone-or-encoder_SA_SS})
    & \cellcolor{mytabcolor_1}\checkmark
    & \cellcolor{mytabcolor_1} 
    & \cellcolor{mytabcolor_1}
    & \cellcolor{mytabcolor_1}
    & \cellcolor{mytabcolor_1}
    & \cellcolor{mytabcolor_1}\checkmark
    & \cellcolor{mytabcolor_1}
    & \cellcolor{mytabcolor_1}
    & \cellcolor{mytabcolor_1}
    \\ 
    
    ~
    & 
    & MS-CSDG (\S\ref{sec_DG_CLIP-as-backbone-or-encoder_SA_MS_CSDG})
    & 
    & \checkmark
    & 
    & 
    & 
    & \checkmark
    & 
    & 
    \\ 
    
    ~
    & 
    & \cellcolor{mytabcolor_1}{MS-OSDG (\S\ref{sec_DG_CLIP-as-backbone-or-encoder_SA_MS_OSDG})}
    & \cellcolor{mytabcolor_1}
    & \cellcolor{mytabcolor_1}\checkmark
    & \cellcolor{mytabcolor_1}
    & \cellcolor{mytabcolor_1}
    & \cellcolor{mytabcolor_1}
    & \cellcolor{mytabcolor_1}
    & \cellcolor{mytabcolor_1}
    & \cellcolor{mytabcolor_1}\checkmark
    & \cellcolor{mytabcolor_1}
    \\ 
    \cmidrule(lr){2-2}
    
    ~
    & Source-Free (SF)
    & S(F)F-DG (\S\ref{sec_DG_CLIP-as-backbone-or-encoder_SF})
    & 
    & 
    & 
    & \checkmark
    & 
    & 
    & 
    & 
    & 
    \\ 
    \midrule
    
    \multirow{10}{*}{\makecell{Domain \\Adaptation\\(DA)}}
    & \multirow{4}{*}{\makecell{Source-Available (SA)}}
    & \cellcolor{mytabcolor_3}{SS-CSUDA (\S\ref{sec_DA_SA_SS_CS})}
    & \cellcolor{mytabcolor_3}\checkmark
    & \cellcolor{mytabcolor_3}
    & \cellcolor{mytabcolor_3}\checkmark
    & \cellcolor{mytabcolor_3}\checkmark
    & \cellcolor{mytabcolor_3}\checkmark
    & \cellcolor{mytabcolor_3}\checkmark
    & \cellcolor{mytabcolor_3}
    & \cellcolor{mytabcolor_3}
    & \cellcolor{mytabcolor_3}
    \\ 
    
    ~
    & 
    & SS-OSUDA (\S\ref{sec_DA_SA_SS_OS})
    & \checkmark
    & 
    & \checkmark
    & \checkmark
    &  \checkmark
    & 
    & 
    & \checkmark
    & 
    \\ 
    
    ~
    & 
    & \cellcolor{mytabcolor_3}{MS-CSUDA (\S\ref{sec_DA_SA_MS_CS})}
    & \cellcolor{mytabcolor_3}
    & \cellcolor{mytabcolor_3}\checkmark
    & \cellcolor{mytabcolor_3}\checkmark
    & \cellcolor{mytabcolor_3}\checkmark
    & \cellcolor{mytabcolor_3}\checkmark
    & \cellcolor{mytabcolor_3}\checkmark
    & \cellcolor{mytabcolor_3}
    & \cellcolor{mytabcolor_3}
    & \cellcolor{mytabcolor_3}
    \\

    ~
    & 
    & MS-OPSUDA (\S\ref{sec_DA_SA_MS_OPS})
    & 
    & \checkmark
    & \checkmark
    & \checkmark
    & \checkmark
    & 
    & 
    & 
    & \checkmark
    \\

    ~
    & \multirow{7}{*}{\makecell{Source-Fre (SF)}}
    & \cellcolor{mytabcolor_4}{SFF-CSUDA a.k.a. CS-UFT (\S\ref{sec_DA_SF_SFF_CS})}
    & \cellcolor{mytabcolor_4}
    & \cellcolor{mytabcolor_4}
    & \cellcolor{mytabcolor_4}
    & \cellcolor{mytabcolor_4}
    & \cellcolor{mytabcolor_4}\checkmark
    & \cellcolor{mytabcolor_4}\checkmark
    & \cellcolor{mytabcolor_4}
    & \cellcolor{mytabcolor_4}
    & \cellcolor{mytabcolor_4}
    \\ 
    
    ~
    & 
    & {SFF-PSUDA a.k.a. PS-UFT (\S\ref{sec_DA_SF_SFF_PS})}
    & 
    & 
    & 
    & 
    & \checkmark
    & 
    & \checkmark
    & 
    & 
    \\ 
    
    ~
    & 
    & \cellcolor{mytabcolor_4}{SFF-OSUDA a.k.a. OS-UFT (\S\ref{sec_DA_SF_SFF_OS})}
    & \cellcolor{mytabcolor_4}
    & \cellcolor{mytabcolor_4}
    & \cellcolor{mytabcolor_4}
    & \cellcolor{mytabcolor_4}
    & \cellcolor{mytabcolor_4}\checkmark
    & \cellcolor{mytabcolor_4}
    & \cellcolor{mytabcolor_4}
    & \cellcolor{mytabcolor_4}\checkmark
    & \cellcolor{mytabcolor_4}
    \\ 
    
    ~
    &
    & SFF-OPSUDA a.k.a. OPS-UFT (\S\ref{sec_DA_SF_SFF_OPS})
    & 
    & 
    & 
    & 
    & \checkmark
    & 
    & 
    & 
    & \checkmark
    \\ 
    
    ~ 
    &  
    & \cellcolor{mytabcolor_4}{SDF-CSUDA (\S\ref{sec_DA_SF_SDF_CS})}
    & \cellcolor{mytabcolor_4}
    & \cellcolor{mytabcolor_4}
    & \cellcolor{mytabcolor_4}
    & \cellcolor{mytabcolor_4}\checkmark
    & \cellcolor{mytabcolor_4}\checkmark
    & \cellcolor{mytabcolor_4}\checkmark
    & \cellcolor{mytabcolor_4}
    & \cellcolor{mytabcolor_4}
    & \cellcolor{mytabcolor_4}
    \\ 
    
    ~
    & 
    & SDF-PSUDA (\S\ref{sec_DA_SF_SDF_PS})
    & 
    & 
    & 
    & \checkmark
    & \checkmark
    & 
    & \checkmark
    & 
    & 
    \\ 
    
    ~
    &
    & \cellcolor{mytabcolor_4}{SDF-OSUDA (\S\ref{sec_DA_SF_SDF_OS})}
    & \cellcolor{mytabcolor_4}
    & \cellcolor{mytabcolor_4}
    & \cellcolor{mytabcolor_4}
    & \cellcolor{mytabcolor_4}\checkmark
    & \cellcolor{mytabcolor_4}\checkmark
    & \cellcolor{mytabcolor_4}
    & \cellcolor{mytabcolor_4}
    & \cellcolor{mytabcolor_4}\checkmark
    & \cellcolor{mytabcolor_4}
    \\ 
    \bottomrule
    
    \end{tabular}
}
\end{table*}

\subsubsection{Source-Data-Free (SDF)}
\label{sec_DA_SF_SDF}

\paragraph{\textbf{Source-Data-Free Closed-Set Unsupervised Domain Adaptation (SDF-CSUDA)}}
\label{sec_DA_SF_SDF_CS}

\textit{In SDF-CSUDA, the label space of the target domain $ \mathcal{Y}_t $ is identical to that of the source domain $ \mathcal{Y}_s $, i.e., $ \mathcal{Y}_t = \mathcal{Y}_s $. In this scenario, a model $ f: \mathbb{R}^{H \times W} \mapsto \mathbb{R}^{C} $ is adapted to a target domain $ \mathcal{D}_t = \{x_j\}_{j=1}^{N_t} $ consisting of unlabeled samples, with no access to data from the source domain $ \mathcal{D}_s $, 
but access to source domain knowledge (e.g., a pretrained model on the source domain).
}

Existing studies exploit pretrained source knowledge (e.g., model weights or stored statistics) to guide adaptation when source data is unavailable. Methods typically refine pseudo labels, calibrate prediction confidence, and distill reliable supervision from vision–language priors, enabling effective closed-set alignment under data-free constraints.

DIFO~\cite{2024_CVPR_DIFO} introduces a distillation approach for multimodal models, combining prompt learning and knowledge distillation with regularization for improved reliability.
Similarly,
Co-learn++~\cite{2024_IJCV_Co-learn++} enhances target adaptation by integrating pre-trained networks. It improves pseudolabel quality through collaboration with the source model and feature extractor, leveraging CLIP's zero-shot classification decisions.
In addition,
CDBN~\cite{2024_arXiv_CDBN} integrates source domain class semantics into unsupervised fine-tuning for target domain generalization, preserving class information while enhancing accuracy and diversity in predictions.
Moreover,
\citet{2024_Signal-Image-and-Video-Processing_BBC} introduces BBC, a black-box domain adaptation method using CLIP. It improves pseudo-label accuracy through joint label generation from a cloud API and CLIP, along with a structure-preserved strategy refining labels using k-nearest neighbors.
Further,
ProDe~\cite{2025_ICLR_ProDe_Proxy-Denoising-for-Source-Free-Domain-Adaptation} introduces a proxy denoising framework that corrects noisy VLM predictions through a reliable proxy. It models the dynamic effect of proxy’s divergence against the latent domain-invariant space and employs mutual knowledge distilling regularization.
Notably,
DUET~\cite{2025_NeurIPS_DUET_DUET=Dual-Perspective-Pseudo-Labeling-and-Uncertainty-aware-Exploration-Exploitation-Training-for-Source-Free-Domain-Adaptation} integrates task-specific and domain-invariant predictions through dual-perspective pseudo labeling, combined with a Tsallis mutual information–based uncertainty-aware optimization that dynamically balances exploration and exploitation during CLIP vision encoder fine-tuning.

\paragraph{\textbf{Source-Data-Free Partial-Set Unsupervised Domain Adaptation (SDF-PSUDA)}}
\label{sec_DA_SF_SDF_PS}

\textit{In SDF-PSUDA, the label space of target domain $ \mathcal{Y}_t $ is a subset of the label space of source domain $ \mathcal{Y}_s $, i.e., $ \mathcal{Y}_t \subset \mathcal{Y}_s $. In this scenario, a model $ f: \mathbb{R}^{H \times W} \mapsto \mathbb{R}^{C} $ is adapted to a target domain $ \mathcal{D}_t = \{x_j\}_{j=1}^{N_t} $ consisting of unlabeled samples, with no access to data from the source domain $ \mathcal{D}_s $, 
but access to source domain knowledge (e.g., a pretrained model on the source domain).
}

Existing studies primarily address this scenario by refining pseudo labels, filtering irrelevant categories using vision–language priors, and stabilizing learning through mutual-information or proxy-denoising mechanisms, enabling CLIP to adapt effectively when only a subset of source classes appears in the target domain.

In this context, DIFO~\cite{2024_CVPR_DIFO} employs mutual-information maximization to refine class-relevant signals within the partial label space, enabling accurate adaptation to the target class distribution without any labeled source data.
Colearn++~\cite{2024_IJCV_Co-learn++} also demonstrates strong performance in this scenario, effectively adapting to target domains with a subset of source classes and improving classification accuracy.
In parallel, 
ProDe~\cite{2025_ICLR_ProDe_Proxy-Denoising-for-Source-Free-Domain-Adaptation} maintains reliable VLM guidance by filtering irrelevant categories through proxy confidence estimation, effectively reducing negative transfer from non-overlapping classes.

\paragraph{\textbf{Source-Data-Free Open-Set Unsupervised Domain Adaptation (SDF-OSUDA)}}
\label{sec_DA_SF_SDF_OS}

\textit{In SDF-OSUDA, the label space of the source domain $ \mathcal{Y}_s $ is a subset of the label space of the target domain $ \mathcal{Y}_t $, i.e., $ \mathcal{Y}_s \subset \mathcal{Y}_t $. In this context, a model $ f: \mathbb{R}^{H \times W} \mapsto \mathbb{R}^{C} $ is adapted to a target domain $ \mathcal{D}_t = \{x_j\}_{j=1}^{N_t} $ that contains unlabeled samples, with no access to data from the source domain $ \mathcal{D}_s $, 
but access to source domain knowledge (e.g., a pretrained model on the source domain).
}

Existing studies in this scenario primarily strengthen open-set robustness by refining CLIP-based supervision, distilling transferable semantics, and incorporating confidence-aware mechanisms to detect unknown classes while preserving reliable classification of known ones.

ODA with CLIP~\cite{2023_arXiv_2025_CVIU_ODA-with-CLIP} investigates adaptable methods for leveraging CLIP, assuming access to a model pre-trained on source domain samples and directly using its parameters to initialize the model for the target domain.
Meanwhile,
DIFO~\cite{2024_CVPR_DIFO} distills knowledge from the customized VLM to the target model, enhancing generalization and novel-class recognition while maintaining robust open-set performance and effectively leveraging unlabeled target data.
Moreover,
\citet{2024_IJCV_Co-learn++} further evaluates this scenario, showing effective identification of unseen classes while maintaining robust generalization in open-set scenarios.
Recently, 
ProDe~\cite{2025_ICLR_ProDe_Proxy-Denoising-for-Source-Free-Domain-Adaptation} adopts adaptive proxy denoising to correct VLM supervision and separate unknown samples.

\begin{table*}[t!]
\centering
\caption{\textcolor{revision-color_1}{Common datasets in DG and DA. (\#: number. The statistics follow the commonly adopted versions in recent studies, as several datasets have been reprocessed since their original release.)}}

\label{tab:Datasets}
\renewcommand{\arraystretch}{1.4}
\scalebox{0.62}{
    \begin{tabular}{ >{\centering\arraybackslash}p{5cm} 
                    >{\centering\arraybackslash}p{1.5cm} 
                    >{\centering\arraybackslash}p{3cm} 
                    >{\centering\arraybackslash}p{17cm}
                    }
    \toprule
    \textbf{Dataset} 
        & \makecell{\textbf{\#Domain}} 
        & \makecell{\textbf{\#Category}}  
        & \textbf{\makecell[c]{Description}} 
    \\
    \midrule

    \cellcolor{mytabcolor_3}Office-Home~\cite{2017_CVPR_Office-Home-dataset_Deep-Hashing-Network-for-Unsupervised-Domain-Adaptation} 
        & \cellcolor{mytabcolor_3}4 
        & \cellcolor{mytabcolor_3}65 
        & \cellcolor{mytabcolor_3} Office and home objects from Art, Clipart, Product, and Real-World domains with rich visual diversity.
        \\ 
        
        Office-31~\cite{2010_ECCV_Office-31-dataset_Adapting-Visual-Category-Models-to-New-Domains} 
        & 3 
        & 31 
        & Office environment objects captured from Amazon, DSLR, and Webcam domains under varying conditions.
        \\ 
        
        \cellcolor{mytabcolor_3}VisDA-2017~\cite{2017_arXiv_VisDA-dataset} 
        & \cellcolor{mytabcolor_3}2 
        & \cellcolor{mytabcolor_3}12 
        & \cellcolor{mytabcolor_3}Synthetic-to-real visual dataset containing diverse 2D object categories across multiple visual styles.
        \\ 
        
        DomainNet~\cite{2019_ICCV_DomainNet-dataset_Moment-Matching-for-Multi-source-Domain-Adaptation} 
        & 6 
        & 345
        & Large-scale dataset including Clipart, Real, Sketch, Painting, Infograph, and Quickdraw domains.
        \\ 
        
        \cellcolor{mytabcolor_3}PACS~\cite{2017_ICCV_PACS-dataset_Deep-Broader-and-Artier-Domain-Generalization} 
        & \cellcolor{mytabcolor_3}4 
        & \cellcolor{mytabcolor_3}7 
        & \cellcolor{mytabcolor_3}Photo, Art, Cartoon, and Sketch domains containing diverse objects with significant style variation.
        \\ 
        
        VLCS~\cite{2013_ICCV_VLCS-dataset_Unbiased-Metric-Learning--On-the-Utilization-of-Multiple-Datasets-and-Web-Images-for-Softening-Bias} 
        & 4 
        & 5 
        & Multi-domain dataset combining VOC2007, LabelMe, Caltech101, and SUN09 with common object types.
        \\ 
        
        \cellcolor{mytabcolor_3}Digits-DG~\cite{2020_ECCV_Digits-DG-dataset_Learning-to-Generate-Novel-Domains-for-Domain-Generalization} 
        & \cellcolor{mytabcolor_3}4 
        & \cellcolor{mytabcolor_3}10 
        & \cellcolor{mytabcolor_3}Handwritten digits from MNIST, MNIST-M, SVHN, and SYNTH datasets covering multiple visual sources.
        \\

        TerraIncognita~\cite{2018_ECCV_Recognition-in-Terra-Incognita}
        & 4 
        & 10 
        & Wildlife camera trap images from different geographic regions with various environmental contexts.
        \\ 
        
        \cellcolor{mytabcolor_3}NICO++~\cite{2023_CVPR_NICO++} 
        & \cellcolor{mytabcolor_3}6
        & \cellcolor{mytabcolor_3}60
        & \cellcolor{mytabcolor_3}Contextual recognition dataset featuring objects with diverse backgrounds and contextual variations.
        \\
        \midrule

        ImageNet~\cite{2009_CVPR_ImageNet}
        & 1 
        & 1000
        & Large-scale natural image dataset covering a wide variety of everyday object categories.
        \\ 
        
        \cellcolor{mytabcolor_4}ImageNetV2~\cite{2019_ICML_ImageNetV2-dataset_Do-ImageNet-Classifiers-Generalize-to-ImageNet} 
        & \cellcolor{mytabcolor_4}1
        & \cellcolor{mytabcolor_4}1000
        & \cellcolor{mytabcolor_4}Re-collected version of ImageNet containing matched categories under new sampling distributions.
        \\ 
        
        ImageNet-Sketch~\cite{2019_NeurIPS_ImageNet-Sketch-dataset_Learning-Robust-Global-Representationis-by-Penalizing-Local-Predictive-Power} 
        & 1 
        & 1000 
        & Sketch drawings of ImageNet categories representing shape-based depictions of natural objects.
        \\ 
        
        \cellcolor{mytabcolor_4}ImageNet-A~\cite{2021_CVPR_ImageNet-A-dataset_Natural-Adversarial-Examples} 
        & \cellcolor{mytabcolor_4}1 
        & \cellcolor{mytabcolor_4}200 
        & \cellcolor{mytabcolor_4}Adversarial natural images designed to induce misclassification in ImageNet-trained models.
        \\ 
        
        ImageNet-R~\cite{2021_ICCV_ImageNet-R-dataset_The-Many-Faces-of-Robustness--A-Critical-Analysis-of-Out-of-Distribution-Generalization} 
        & 1 
        & 200 
        & Rendered images including art, cartoon, sculpture, and painting styles derived from ImageNet classes.
        \\ 
        
        \cellcolor{mytabcolor_4}CIFAR-10~\cite{2009_Toronto-on-Canada_CIFAR10-and-CIFAR100-dataset_Learning-Multiple-Layers-of-Features-from-Tiny-Images} 
        & \cellcolor{mytabcolor_4}1 
        & \cellcolor{mytabcolor_4}10 
        & \cellcolor{mytabcolor_4}Small-scale natural image dataset of everyday objects such as animals, vehicles, and tools.
        \\
        
        CIFAR-100~\cite{2009_Toronto-on-Canada_CIFAR10-and-CIFAR100-dataset_Learning-Multiple-Layers-of-Features-from-Tiny-Images} 
        & 1 
        & 100 
        & Expanded version of CIFAR10 with 100 fine-grained and hierarchically structured
        \\
        
        \cellcolor{mytabcolor_4}Caltech101~\cite{2004_CVPRW_Caltech101-dataset_Learning-Generative-Visual-Models-from-Few-Training-Examples} 
        & \cellcolor{mytabcolor_4}1 
        & \cellcolor{mytabcolor_4}100
        & \cellcolor{mytabcolor_4}Object categories including faces, animals, plants, and household items captured in controlled settings.
        \\
        
        DTD~\cite{2014_CVPR_DTD-dataset_Describing-Textures-in-the-Wild} 
        & 1 
        & 47 
        & Texture dataset describing perceptual attributes such as “bumpy,” “striped,” and “woven” patterns.
        \\
        
        \cellcolor{mytabcolor_4}EuroSAT~\cite{2019_J-STAR_EuroSAT-dataset} 
        & \cellcolor{mytabcolor_4}1 
        & \cellcolor{mytabcolor_4}10 
        & \cellcolor{mytabcolor_4}Satellite imagery dataset showing different land use and land cover types across Europe.
        \\
        
        FGVC-Aircraft~\cite{2013_arXiv_FGCV-Aircraft-dataset_Fine-grained-Visual-Classification-of-Aircraft} 
        & 1 
        & 100 
        & Fine-grained collection of aircraft images categorized by model, manufacturer, and configuration.
        \\
        
        \cellcolor{mytabcolor_4}Food101~\cite{2014_ECCV_Food101-dataset} 
        & \cellcolor{mytabcolor_4}1 
        & \cellcolor{mytabcolor_4}101 
        & \cellcolor{mytabcolor_4}Dataset of one 101 food dish categories collected under diverse photographic styles.
        \\
        
        Flowers102~\cite{2009_ICCV_Flower103-dataset_Automated-Flower-Classification-over-A-Large-Number-of-Classes} 
        & 1 
        & 102
        & Flower species of varied shapes, colors, and petal textures collected in multiple lighting conditions.
        \\
        
        \cellcolor{mytabcolor_4}OxfordPets~\cite{2012_CVPR_OxfordPets-dataset_Cats-and-Dogs} 
        & \cellcolor{mytabcolor_4}1 
        & \cellcolor{mytabcolor_4}37
        & \cellcolor{mytabcolor_4}Pet breeds including cats and dogs captured with varied poses, colors, and surrounding contexts.
        \\
        
        SUN397~\cite{2010_CVPR_SUN-Database} 
        & 1 
        & 397 
        & Large-scale scene dataset covering a broad range of indoor and outdoor environment categories.
        \\
        
        \cellcolor{mytabcolor_4}StanfordCars~\cite{2013_ICCVW_StandfordCars-dataset_3D-Object-Representations-for-Fine-grained-Categorization} 
        & \cellcolor{mytabcolor_4}1 
        & \cellcolor{mytabcolor_4}196
        & \cellcolor{mytabcolor_4}High-resolution car images from multiple viewpoints and manufacturing years with fine-grained labels.
        \\
        
        UCF101~\cite{2012_arXiv_UCF101-dataset_UCF101=A-dataset-of-101-Human-Actions-Classes-from-Videos-in-the-Wild} 
        & 1 
        & 101 
        & Human action dataset collected from YouTube videos showing diverse motion types and camera angles.
        \\
        \bottomrule
    \end{tabular}
}

\end{table*}

\subsection{Discussion}
\label{sec_DA_Discussion}

\textcolor{revision-color_1}{
Domain adaptation (DA) has witnessed substantial progress, with a growing number of CLIP-based approaches proposed in recent years.
To analyze the current status and emerging trends of this field, we focus on the widely adopted Office-Home~\cite{2017_CVPR_Office-Home-dataset_Deep-Hashing-Network-for-Unsupervised-Domain-Adaptation} benchmark, which serves as a representative dataset for evaluating cross-domain adaptation.
Based on the analysis throughout this section and the comparative results summarized in Table~\ref{tab:DA-performance}, several fundamental observations and conclusion can be drawn:
}

\textbf{\textit{(i) Diminishing reliance on source data.}} On ResNet-50, zero-shot CLIP averages 71.8, while SDF methods already approach or surpass most source-dependent ones; ProDe \cite{2025_ICLR_ProDe_Proxy-Denoising-for-Source-Free-Domain-Adaptation} reaches 81.1, a gain of 9.3. Within ViT-B/16, SDF with CDBN \cite{2024_arXiv_CDBN} attains 86.2, only 2.2 points behind the best source-dependent method CMKD \cite{2024_TCSVT_CMKD} at 88.4. SFF-UDA methods also advance the frontier; ReCLIP \cite{2024_WACV_ReCLIP} achieves 87.0 on average and surpasses most SA-UDA and SDF-UDA approaches, reinforcing the reduced dependence on source data.

\textbf{\textit{(ii) Prompt tuning dominates yet remains sensitivity-prone.}} Performance varies more on difficult transfers, especially toward the Clipart domain \cite{2017_CVPR_Office-Home-dataset_Deep-Hashing-Network-for-Unsupervised-Domain-Adaptation}. Methods emphasizing alignment distillation and consistency are more stable under large domain shifts; for example, CMKD \cite{2024_TCSVT_CMKD} reports 81.2 from Real to Clipart and 79.4 from Art to Clipart.

\textbf{\textit{(iii) Transformer encoders generalize better across domains.}} Zero-shot ViT-B/16 reaches 82.4 versus 71.8 for ResNet-50, a margin of 10.6 points. This stronger prior raises the ceiling for both SDF and source-dependent methods and enables more flexible prompt and alignment designs.

\textbf{\textit{Overall.}} CLIP-driven DA is moving from \textit{data-driven adaptation} to \textit{knowledge- and prior-driven alignment}. The gap between SDF/SFF and source-dependent methods continues to narrow, while transferring under large domain discrepancy, open-domain robustness, and semantic consistency remain open challenges.

\noindent\colorbox{cyan!15}{\textbf{\textit{Scenarios Summary.}}}
After covering both CLIP-powered DG~(\S\ref{sec_DG}) and DA~(\S\ref{sec_DA}), Table~\ref{tab:scenario-DG-DA} provides an integrated summary of their scenario taxonomy, connecting all settings through source accessibility (SA vs. SF), source domain number (SS vs. MS), and label relationship (CS, PS, OS, OPS), enabling a clearer comparison across different configurations.

\section{Benchmarks and Metrics}
\label{sec_Benchmarks-and-Metrics}

This section introduces common benchmarks (\S\ref{sec_Common-Benchmarks}) and evaluation metrics (\S\ref{sec_Common-Metrics}) for DG and DA. Datasets are categorized into multi-domain and single-domain types, while metrics such as accuracy and harmonic mean assess performance, particularly when unknown target classes are involved. These benchmarks and metrics form the basis for comparing the effectiveness of DA and DG methods.

\subsection{Common Bechmarks}
\label{sec_Common-Benchmarks}

\subsubsection{Datasets for DG and DA}

DA and DG both use datasets that can be categorized into multi-domain and single-domain types, as shown in Table~\ref{tab:Datasets}. In a multidomain dataset, each subset represents a distinct domain with the same categories but different image distributions. In DA and DG experiments, one or more domains act as the source domains (i.e., single-source~\cite{2023_ICCV_PADCLIP, 2023_ICCVW_AD-CLIP, 2024_CVPR_DAMP} or multi-source~\cite{2023_NeurIPS_MPA, 2024_arXiv_LanDA, 2024_OpenReview_MSDPL}) while the remaining domains serve as target domains to evaluate model performance. The domain and category configurations of each dataset follow those used in several recent representative studies~\cite{2020_arxiv_domainbed,2025_IJCAI_BOLD,2022_IJCV_CoOp,2023_CVPR_KgCoOp,2023_ICCV_PADCLIP}.

The primary difference between DA and DG lies in their respective training and evaluation approaches: DA~\cite{2023_ICCV_PADCLIP, 2023_TNNLS_DAPrompt, 2024_CVPR_UniMoS} typically uses labeled source domains and unlabeled target domains, which are jointly trained before testing in the target domain, while DG~\cite{2023_TJSAI_DPL, 2024_WACV_StyLIP, 2024_arXiv_SPG} involves training on one or more labeled source domains, and, after source training, testing is conducted directly in the target domain without additional adaptation.

Single-domain datasets, used in SFF-DA methods~\cite{2022_arXiv_UPL, 2024_WACV_ReCLIP}, leverage CLIP’s zero-shot capabilities for unsupervised training and adaptation. Datasets like ImageNet~\cite{2009_CVPR_ImageNet} and its variants(e.g., ImageNetV2~\cite{2019_ICML_ImageNetV2-dataset_Do-ImageNet-Classifiers-Generalize-to-ImageNet}, ImageNet-Sketch~\cite{2019_NeurIPS_ImageNet-Sketch-dataset_Learning-Robust-Global-Representationis-by-Penalizing-Local-Predictive-Power}, ImageNet-A~\cite{2021_CVPR_ImageNet-A-dataset_Natural-Adversarial-Examples}, and ImageNet-R~\cite{2021_ICCV_ImageNet-R-dataset_The-Many-Faces-of-Robustness--A-Critical-Analysis-of-Out-of-Distribution-Generalization}) serve as benchmarks for evaluating domain generalization in few-shot prompt learning.

\subsection{Common Metrics}
\label{sec_Common-Metrics}

Both DA and DG methods evaluate performance in the target domain. For DA, metrics depend on the category shift between source set $ \mathcal{Y}_s $ and target set $ \mathcal{Y}_t $. For DG, metrics are based on the shift between predefined $ \mathcal{Y}_s $ and the actual target set $ \mathcal{Y}_t $. Specifically:

1. For $ \mathcal{Y}_s = \mathcal{Y}_t $ or $ \mathcal{Y}_t \subset \mathcal{Y}_s $ with $ \mathcal{Y}_s \setminus \mathcal{Y}_t \neq \emptyset $ cases (i.e., closed-set and partial-set scenarios): As there are no unknown classes in the target domain, prior methods typically use either the average class accuracy (ACC) or domain average accuracy across all transfer tasks to evaluate performance.

2. For $ \mathcal{Y}_s \subset \mathcal{Y}_t $ with $ \mathcal{Y}_t \setminus \mathcal{Y}_s \neq \emptyset $ or $ \mathcal{Y}_s \cap \mathcal{Y}_t \neq \emptyset $, $ \mathcal{Y}_t \nsubseteq \mathcal{Y}_s $, with $ \mathcal{Y}_s \nsubseteq \mathcal{Y}_t $ cases (i.e., open-set and open-partial-set scenarios): During target domain evaluation, we commonly define $ \mathcal{Y}_{shared} = \mathcal{Y}_s \cap \mathcal{Y}_t $ as the set of shared classes. Samples within $ \mathcal{Y}_{shared} $ in the target domain are considered known classes, while others are treated as unknown classes. All unknown classes are grouped into a unified "unknown" category. For evaluation, the model must identify these unknown class samples, and performance is measured using the harmonic mean of the known class accuracy (ACC) and unknown class detection accuracy (AUC), calculated as 

\begin{equation}
    HOS = \frac{2 \times ACC_{know} \times ACC_{unknow}}{ACC_{know} + ACC_{unknow}}.
\end{equation}

Notably, some methods use per-class accuracy for $ ACC_{know} $ (e.g., UEO~\cite{2024_ICML_UEO}, DIFO~\cite{2024_CVPR_DIFO}), while others calculate accuracy across all known class samples (e.g., ODA with CLIP~\cite{2023_arXiv_2025_CVIU_ODA-with-CLIP}).

\section{Challenges and Future Directions}
\label{sec_Challenges-and-Future-Directions}

As CLIP-based DG and DA advance, research is moving toward more realistic and multimodal settings with incomplete labels and restricted source availability. Progress is evident, yet challenges in generalization, knowledge usage, interpretability, and stability persist. Addressing these issues is essential for developing scalable and reliable CLIP-powered systems.

\textbf{Evolving toward Realistic Scenarios.}  
Recent research trends show a gradual shift from source-available (SA) and closed-set (CS) assumptions~\cite{2023_ICCV_BorLan, 2023_ICCV_RISE, 2023_TNNLS_DAPrompt, 2023_NeurIPS_MPA} toward more realistic and efficient settings (see Fig.~\ref{fig:Roadmap} and Table~\ref{tab:scenario-DG-DA}), such as source-free (SF)~\cite{2023_ICML_POUF, 2024_NeurIPS_LaFTer}, partial-set (PS)~\cite{2024_ICML_UEO, 2024_CVPR_DIFO,2024_IJCV_Co-learn++, 2025_ICLR_ProDe_Proxy-Denoising-for-Source-Free-Domain-Adaptation}, open-set (OS)~\cite{2024_ICIP_PromptDIV, 2023_ICML-Workshop_UOTA}, and open-partial-set (OPS)~\cite{2024_Signal-Processing-Letters_SAP-CLIP, 2024_ICML_UEO}, which better reflect real-world deployment where labeled source data are inaccessible and category overlaps incomplete. Such transitions mark an important step for CLIP-based DG and DA toward practical generalization, emphasizing efficiency, scalability, and adaptability under realistic constraints,
while also raising new challenges in handling uncertainty, long-tailed categories, and deployment-time robustness.

\textbf{Incorporating LLM Knowledge.}
Although CLIP demonstrates strong vision–language alignment, its inherent knowledge remains limited to paired image–text supervision and lacks broader contextual reasoning. 
Recent studies~\cite{2025_CVPR_OSLoPrompt_OSLoPrompt=Bridging-Low-Supervision-Challenges-and-Open-Set-Domain-Generalization-in-CLIP, 2025_TCSVT_BatStyler_BatStyler=Advancing-Multi-category-Style-Generation-for-Source-free-Domain-Generalization, 2025_IJCV_MAwLLM_Multi-modal-Prompt-Alignment-with-Fine-grained-LLM-Knowledge-for-Unsupervised-Domain-Adaptation} suggest that integrating knowledge from large language models (LLMs) can enhance domain reasoning and open-domain adaptability while providing richer semantic grounding. 
Future research should explore efficient strategies for injecting LLM (e.g., GPT-4~\cite{2023_arXiv_GPT-4_GPT-4-technical-Report}, Gemini~\cite{2023_arXiv_Gemini_Gemini=A-Family-of-Highly-Capable-Multimodal-Models}, LLaMA~\cite{2023_arXiv_LLaMA_LLaMA=Open-and-Efficient-Foundation-Language-Models}) knowledge into CLIP-based DG and DA, achieving a balance between enhanced generalization and computational efficiency,
with particular attention to hallucination control, alignment reliability, and inference-time efficiency.

\textbf{Integration of Multimodal Data.} 
Existing CLIP-based DG and DA studies primarily focus on image–text modalities, while real-world applications often involve additional signals such as video~\cite{2023_ICCV_AutoLabel_AutoLabel=CLIP-based-Framework-for-Open-set-Video-Domain-Adaptation, 2023_ICCV_DALL-V_The-Unreasonable-Effectiveness-of-Large-Language-Vision-Models-for-Source-free-Video-Domain-Adaptation, 2024_CVPR_UNITE_Unsupervised-Video-Domain-Adaptation-with-Masked-Pre-training-and-Collaborative-Self-training} and sensor data~\cite{2021_ACM-TOMM_Survey_Survey-on-Deel-Multi-modal-Data-Analytics=Collaboration-Rivalry-and-Fusion}. 
Integrating these modalities can enhance model robustness~\cite{2021_NeurIPS_What-Makes-Multi-modal-Learning-Better-than-Single(Provably)} by providing complementary cues across domains. 
Future architectures should focus on unified multimodal fusion frameworks that can dynamically weight different modalities to improve domain adaptability and semantic alignment under distribution shifts, explicitly addressing challenges such as modality imbalance, noisy observations, and cross-modal consistency.

\textbf{Interpretable CLIP-powered Model.} 
CLIP-powered models often lack interpretability~\cite{2020_arXiv_Survey_Opportunities-and-Challenges-in-Explainable-AI, 2024_DMKD_Survey_A-Comprehensive-Taxonomy-for-Explainable-AI}, making it difficult to understand their decisions, especially in critical domains like healthcare and finance where transparency is vital~\cite{2024_TIST_Survey_Explainability-for-LLMs}. As CLIP-based DG and DA are increasingly deployed in such scenarios, improving interpretability is essential for trustworthy behavior. Future work may leverage explainable AI techniques~\cite{2024_DMKD_Survey_A-Comprehensive-Taxonomy-for-Explainable-AI}, including attention visualization, feature attribution, and semantic reasoning, to enhance transparency and foster human AI collaboration by emphasizing explanation faithfulness, cross-domain consistency, and user-level interpretability.

\textbf{Catastrophic Forgetting.} 
Fine-tuning CLIP-powered models for new domains often leads to catastrophic forgetting~\cite{2018_Lifelong-Machine-Learning_Survey_Continual-Learning-and-Catastrophic-Forgetting, 2018_AAAI_Survey_Measuring-Catastrophic-Forgetting-in-Neural-Networks, 2023_arXiv_Survey_Catastrophic-Forgetting-in-Deel-Learning=A-Comprehensive-Taxonomy}, where previously learned knowledge is overwritten during adaptation. Traditional strategies rely on regularization or rehearsal to retain prior information, while recent methods integrate memory-augmented modules and parameter-isolation mechanisms~\cite{2021_TPAMI_Survey_A-Continual-Learning-Survey, 2024_arXiv_Survey_Continual-Learning-and-Catastrophic-Forgetting}. In CLIP-based DG and DA, combining continual learning with adaptive learning strategies offers a promising path to balance retention and adaptation, ensuring stability across evolving domains and enabling more reliable knowledge transfer under long-term and incremental domain shifts.

\begin{figure}[t]
    \centering
    \includegraphics[width=0.9\linewidth]{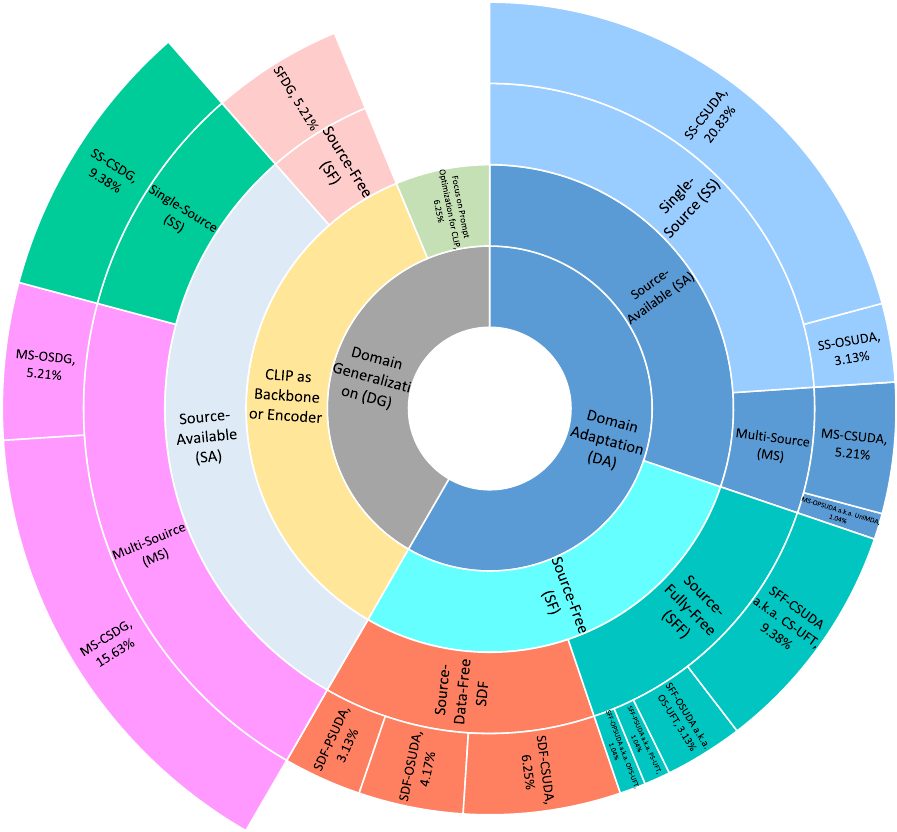}
    \caption{\textcolor{revision-color_1}{Percentage of cited papers in different sections, showing active exploration across scenarios.}}
    \label{fig:TreeMap}
\end{figure}

\section{Conclusion}
\label{sec_Conclusion}
This survey provides an overview of CLIP's applications in domain generalization (DG) and domain adaptation (DA), emphasizing its zero-shot capabilities for handling unseen domains without extensive retraining. It summarizes representative methodologies and synthesizes their progress across different research directions. Fig.~\ref{fig:TreeMap} visualizes the distribution of CLIP-based approaches, revealing their diversity and evolution. Building on the discussions in previous sections, we outline key challenges and potential avenues for future research.

{\footnotesize
\bibliographystyle{plainnat}
\bibliography{Survey_short-version}
}

\end{document}